\documentclass[10pt,twocolumn,letterpaper]{article}

\usepackage[pagenumbers]{iccv} 

\usepackage{marvosym}
\usepackage{amssymb}

\usepackage{algorithm}
\usepackage{algorithmic}
\usepackage{xcolor}
\usepackage{adjustbox}
\usepackage{amsmath} 
\usepackage{amsfonts} 
\usepackage{makecell}
\usepackage{colortbl}
\usepackage{multirow}
\usepackage{booktabs}
\usepackage{bm}
\usepackage{xcolor}

\usepackage{newfloat}
\usepackage{listings}

\definecolor{iccvblue}{rgb}{0.21,0.49,0.74}
\usepackage[pagebackref,breaklinks,colorlinks,allcolors=iccvblue]{hyperref}

\begin{document}

\makeatletter
\def\thanks#1{\protected@xdef\@thanks{\@thanks
        \protect\footnotetext{#1}}}
\makeatother


%

\title{Orthogonal Projection Subspace to Aggregate Online Prior-knowledge for Continual Test-time Adaptation}

\thanks{\textsuperscript{\Letter} Corresponding author.}
\author{Jinlong Li$^{1}$ ~~~~~~~ {Dong Zhao}$^{1}$\textsuperscript{\Letter} ~~~~~~~ {Qi Zang}$^{1}$ ~~~~~~~ {Zequn Jie}$^{2}$ ~~~~~~~ {Lin Ma}$^{2}$  ~~~~~~~  {Nicu Sebe$^{1}$} \\
$^1$ University of Trento \quad  \quad  \quad $^2$ Meituan Inc. \\
}

\newcommand{\jinlong}[1]{\textcolor{violet}{\textbf{Jinlong: }{#1}}}

\definecolor{tabhighlight}{HTML}{e5e5e5}

\maketitle

\begin{abstract}
\label{abstract}

\noindent Continual Test Time Adaptation (CTTA) is a task that requires a source pre-trained model to continually adapt to new scenarios with changing target distributions. Existing CTTA methods primarily focus on mitigating the challenges of catastrophic forgetting and error accumulation. Though there have been emerging methods based on forgetting adaptation with parameter-efficient fine-tuning, they still struggle to balance competitive performance and efficient model adaptation, particularly in complex tasks like semantic segmentation. In this paper, to tackle the above issues, we propose a novel pipeline, \textit{\textbf{\underline O}}rthogonal Projection Subspace to aggregate \textit{\textbf{\underline o}}line \textit{\textbf{\underline P}}rior-\textit{\textbf{\underline k}}nowledge, dubbed \textit{\textbf{OoPk}}. Specifically, we first project a tuning subspace orthogonally which allows the model to adapt to new domains while preserving the knowledge integrity of the pre-trained source model to alleviate catastrophic forgetting. Then, we elaborate an online prior-knowledge aggregation strategy that employs an aggressive yet efficient image masking strategy to mimic potential target dynamism, enhancing the student model's domain adaptability. This further gradually ameliorates the teacher model's knowledge, ensuring high-quality pseudo labels and reducing error accumulation. We demonstrate our method with extensive experiments that surpass previous CTTA methods and achieve competitive performances across various continual TTA benchmarks in semantic segmentation tasks.

\end{abstract}    
\section{Introduction}
\label{sec:intro}

Despite significant advances in deep learning for computer vision~\cite{he2016deep,he2020momentum,he2022masked,dosovitskiy2020image,huang2017densely}, deep neural networks, either CNNs~\cite{he2016deep,huang2017densely} or Transformers~\cite{dosovitskiy2020image,liu2021swin}, fall short of obtaining decent performances when encountering different distributions between source and target domains~\cite{choi2021robustnet,long2016unsupervised,li2021semantic}. To address this, \textit{i.e.,} domain shifts, there has been increasing attention on test-time adaptation (TTA) owing to its practicality and broad applicability, particularly in cutting-edge scenarios~\cite{wang2020tent,liu2021ttt++,iwasawa2021test,gandelsman2022test}. Since the real world usually comes with dynamic environments, deployed models are required to exhibit sufficient generalized capabilities and transferable performances. To this end, TTA methods focus on adapting a pre-trained source model into unknown target domains with unlabeled online data and without access to previous source data, due to privacy concerns or legal constraints.

\begin{figure}[tbp]
    \begin{center}
    \centering 
\includegraphics[width=0.495\textwidth]{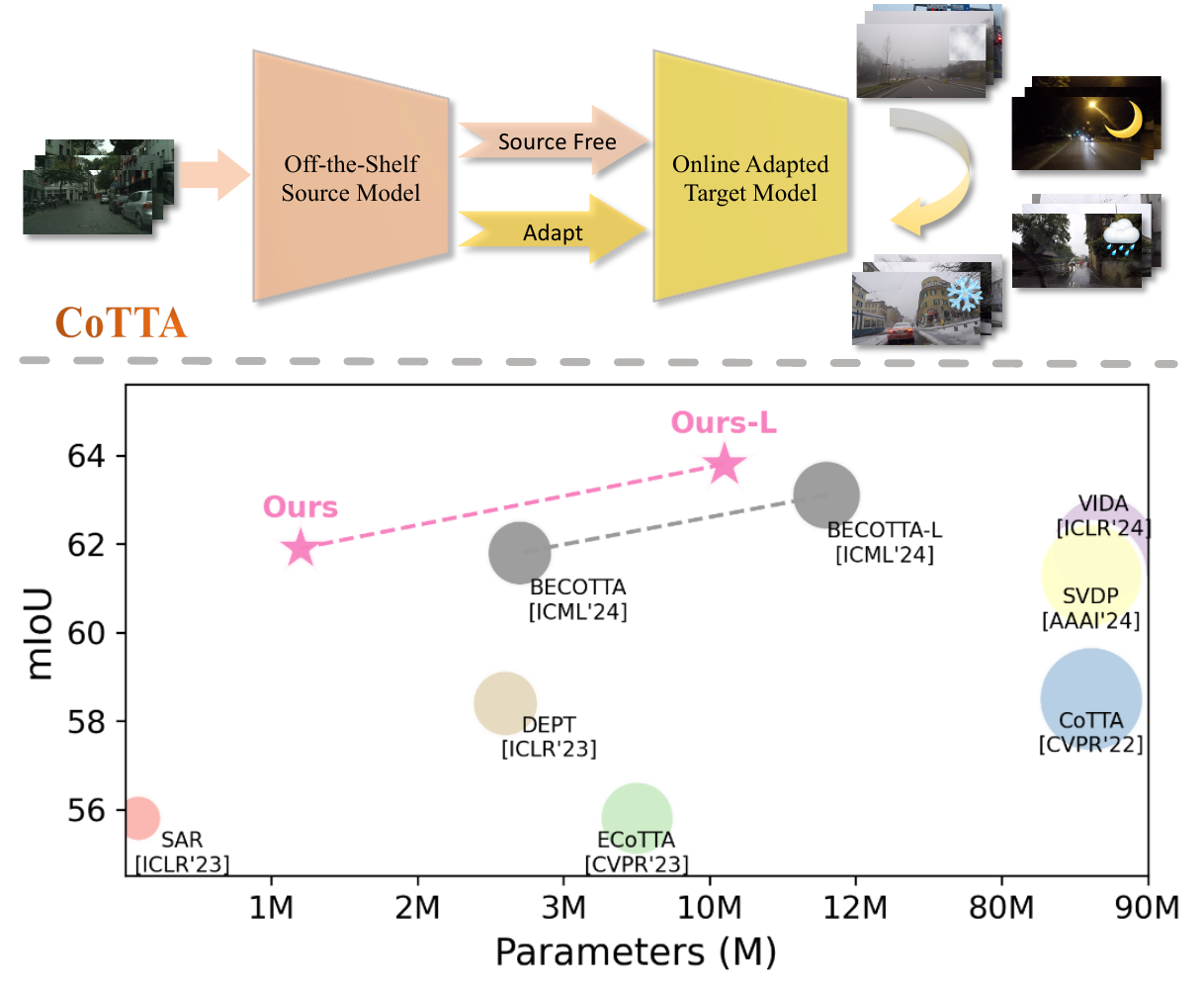}
    \end{center}
    \vspace{-0.30cm}
\setlength{\abovecaptionskip}{-0.05 cm}
    \caption{CoTTA scenario is illustrated above. Bubble chart below compares the training parameters (x-axis) and adaptation performance (y-axis) with other methods. The area of each bubble represents the \textbf{\textit{memory consumption}} during test-time adaptation. It shows that our method is highly competitive in both performance and efficiency.
}
    \label{fig:compares}
\vspace{-0.60cm}
\end{figure}

Whilst the reality is frequently changed due to the complicated environmental scenarios, the problem of continual test-time adaptation (CTTA)~\cite{wang2022continual} has been introduced from TTA with fixed target domain assumption to continuously changing target distributions~\cite{wang2022continual}, aiming at enforcing the model rapidly and continually adapt to newly unseen environments while maintaining relative domain knowledge garnered during adaptation in the long history. 

Solving CTTA problems needs to address two key challenges: \textit{(i)} catastrophic forgetting that gradually forgets previous domain knowledge to deteriorate model's plasticity, and \textit{(ii)} error accumulation that relies on pseudo labels online to realize adaptation which potentially causes inaccurate supervision and constantly degrade model's performance. Meanwhile, deployments are supposed to be embedded into cutting-edge devices, which poses the challenge of efficient adaptation scheme while most CTTA approaches~\cite{wang2022continual,liu2023vida,lee2024becotta} neglect the computational cost to perform heavy teacher-student framework, even with full parameters update.
With the surge of parameter-efficient fine-tuning (PEFT) techniques, such as adapter~\cite{houlsby2019parameter}, LoRA~\cite{hu2022lora} and MoE~\cite{shazeer2017outrageously,fedus2022switch,yang2024exploring}, \textit{etc.} Utility of PEFTs to force the model to adapt to unseen domains has been capturing attention~\cite{lee2024becotta,liu2023vida,yang2024exploring}. Despite the competitive performances obtained, they still necessitate complex training procedures or require Stochastic Restoration to resolve catastrophic forgetting, lacking explicit and essentially efficient adaptation paradigms to solve the CTTA task.

To alleviate these critical issues, we propose a novel pipeline, \textit{\textbf{\underline O}}rthogonal Projection Subspace to aggregate \textit{\textbf{\underline o}}line \textit{\textbf{\underline P}}rior-\textit{\textbf{\underline k}}nowledge, dubbed \textit{\textbf{OoPk}}. Inspired by currently prevalent orthogonal fine-tuning~\cite{Liu2021OPT,LiuNIPS18,qiu2023controlling} for generative models or LLMs, an interesting question arises: shall we fully unleash knowledge from pre-trained source model whilst embracing continuously diverse unseen domain knowledge based on orthogonal transformation, meaning \textit{harmony in diversity}?
The answer is \textbf{\textit{Yes}}. We therefore formulate such domain adaptation into an Orthogonal Projection Subspace (OPS) tuning, achieved by adjusting the coordinate angles of pre-trained source weights to adapt to new domains. Orthogonal fine-tuning intrinsically has norm-preserving properties among neurons, capturing the level of norm uniformity on the unit hypersphere in terms of semantic information encoding, which can be demonstrated via a toy experiment shown in Fig.~\ref{fig:motivation}. 
During the adaptation, the model is gradually adapted to new target domain knowledge within our \textit{orthogonal projection subspace} while still allowing to steer previous domain knowledge.
%
%

Specifically, to ensure efficient and effective adaptation, we employ extra lightweight trainable modules~\cite{hu2022lora,houlsby2019parameter} into the pre-trained source model, keeping the pre-trained source model frozen to reduce training overhead. Then, we apply each additional tuning module with \textit{soft} orthogonal regularization towards identity matrices, like \textit{l2} regularization, that is conducive to preserving unchanged norm-property for capturing new target knowledge into projected subspace and still smoothly acquire useful knowledge from previous domains. Moreover, these additional tuning parameters can be merged into the source model and introduce no inference cost when deploying, which demonstrates promising practicality and broad applicability beyond prompt- and MoE-based methods~\cite{yang2024exploring,lee2024becotta}, comparisons illustrated in Fig.~\ref{fig:compares} and Fig.~\ref{fig:param_compaison}. 
Under the CTTA setting when facing weathers like \textit{fog, rain, night, or snow}, we empirically observe that continuously changing target distribution often comes with serious occlusion, blur, or darkness, \textit{etc}. We further elaborate a novel prior-knowledge aggregation strategy to mimic the potential target domain dynamism better and further induce stronger adaptation learning to strengthen the model's domain adaptability. 
A simple yet effective Image Masking Strategy (IMS) is tailored to maskout image content to simulate challenging target distributions, which progressively enhances the student's target knowledge and ameliorates the teacher's adaptability via the EMA framework. IMS is established for the student model with high-percent random image masking by assigning masked regions with 0/255 proxy pixels, while the teacher model supports pseudo labels for supervising student based on original input.
The student will be \textit{aggressively} forced to fit into new domains and attend more contextual relationships to make predictions given the target samples access online once.

\begin{figure}[tbp]
    \begin{center}
    \centering 
\includegraphics[width=0.48\textwidth]{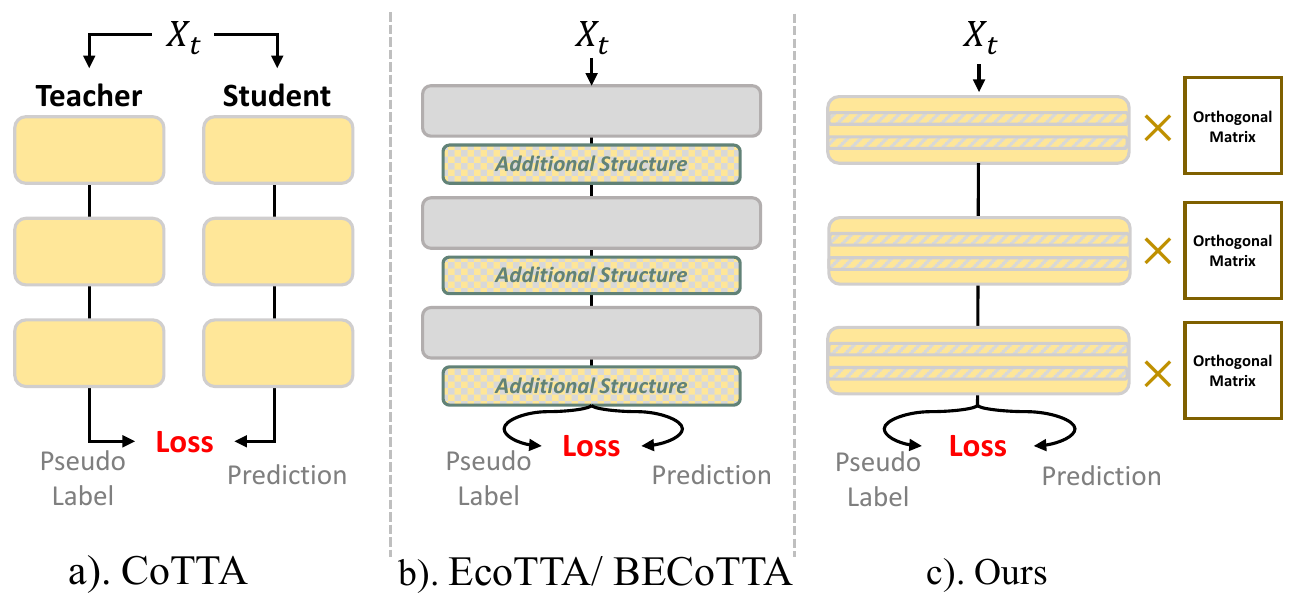}
    \end{center}
    \vspace{-0.20cm}
\setlength{\abovecaptionskip}{-0.05 cm}
    \caption{Principle comparison with CoTTA\cite{wang2022continual}, EcoTTA\cite{song2023ecotta} and BECoTTA\cite{lee2024becotta}. Our method has \textbf{\textit{fewer}} trainable parameters and does not rely on adding new structures via orthogonal projection.
}
    \label{fig:param_compaison}
\vspace{-0.60cm}
\end{figure}

In this way, we bolster the target adaptability of the model by projecting an orthogonal subspace to acquire target domain knowledge through OPS and empower the student-teacher cooperatively to aggregate diverse prior knowledge by IMS within orthogonal projection subspace, striking competitive adaptation performance, particularly for CTTA semantic segmentation task.
In summary, our main contributions are as follows:
\begin{itemize}
    \item We propose a novel adaptation method that projects an orthogonal subspace via lightweight and efficient tuning with soft orthogonal regularization to allow target domain learning while freezing the whole pre-trained source model.
    \item A simple yet effective image masking strategy is proposed to enhance the model's adaptability by aggressively adapting student into new domains, empowering student-teacher cooperatively to aggregate diverse target prior knowledge. 
    \item Extensive experiments demonstrate that our proposed \textit{OoPk} pipeline obtains \textit{SoTA} performances with relatively fewer tunable parameters for CTTA tasks. 
\end{itemize}

\section{Related works}
\label{sec:relatedworks}

\subsection{Test-Time Adaptation}
\vspace{-0.1cm}
Test-time adaptation (TTA)~\cite{boudiaf2023search,kundu2020universal,yang2021generalized,zang2023boosting} is also known as source-free domain adaptation, trying to adapt a pre-trained source model to a changing/unknown target domain distribution without access to any previous source domain data. 
SHOT~\cite{liang2020we} proposes a method by distinguishing itself by focusing on optimizing the backbone exclusively using information maximization and pseudo labels. AdaContrast~\cite{chen2022contrastive} integrate self-supervised contrastive learning methods and also pseudo labels. NOTELA~\cite{boudiaf2023search} manipulates the output distribution to achieve TTA, while ~\cite{wang2020tent,niu2023towards,yuan2023robust} updating the parameters of Batch Normalization layers.

\vspace{-0.1cm}
\subsection{Continual Test Time Adaptation}
\vspace{-0.1cm}
Continual Test Time Adaptation (CTTA) extends the TTA to a more challenging scenario that assumes target domain distributions are dynamic and changing continually. CoTTA~\cite{wang2022continual} first proposes a teacher-student (EMA) pipeline by using consistency loss and stochastic weight reset to mitigate the catastrophic forgotten. EcoTTA~\cite{song2023ecotta} probes meta-networks and self-distillation regularization within an efficient memory consumption. DePT~\cite{gao2022visual} adopts visual prompts~\cite{jia2022visual} to adapt to target distribution. Recently, ViDA~\cite{liu2023vida} and BECoTTA~\cite{lee2024becotta} apply parameter-efficient fine-tuning (PEFT)~\cite{shazeer2017outrageously,fedus2022switch,zuo2021taming} to realize CTTA tasks. While the aforementioned works primarily concentrate on classification tasks, there has been a line of promising works~\cite{yang2024exploring,shin2022mm,song2022cd,zhang2022auxadapt} in applying TTA or CTTA on dense prediction task,~\textit{etc.}

\vspace{-0.1cm}
\subsection{Parameter-Efficient Fine-Tuning}
\vspace{-0.1cm}
With the emergence of VLMs, efficiently adapting models to downstream tasks with limited data samples has garnered significant interest. Parameter-Efficient Fine-Tuning (PEFT)~\cite{lester2021power,li2021prefix,houlsby2019parameter,hu2022lora,he2022towards,gao2024clip,zhang2021tip} emerges, aiming at updating or adding a relatively small number of parameters to the pre-trained model. Adaptation methods~\cite{houlsby2019parameter,hu2022lora,poth-etal-2023-adapters} have become increasingly ubiquitous to efficiently adapt pre-trained models into downstream tasks, while LoRA series~\cite{hu2022lora,dettmers2024qlora} is widely used to fine-tune pre-trained model weights using low-rank matrix optimization.

\vspace{-0.1cm}
\subsection{MAE Image Masking Strategy}
\vspace{-0.1cm}
Among the conventional data augmentation techniques, MixUp~\cite{zhang2017mixup}, CutMix~\cite{yun2019cutmix}, CutOut~\cite{devries2017improved}, Random Erasing~\cite{zhong2020random}, \textit{etc.,} has been widely studied to boost the model's robustness and generalizability. Coming to masked-based self-supervised learning for reconstruction loss driven, a plethora of works~\cite{devlin2018bert,bao2021beit,he2022masked,xie2022simmim} have been successfully reducing the large-scale annotation cost in both Natural Language Processing and Computer Vision.
, involving randomly masking out a significant percentage of input patch tokens. 
Differently, in this paper, we employ a masking strategy to enforce model adapt to learn better target prior knowledge instead of reconstruction.

\section{Methodology}
\label{sec:method}

\begin{figure*}[t]
\includegraphics[width=0.99\textwidth]{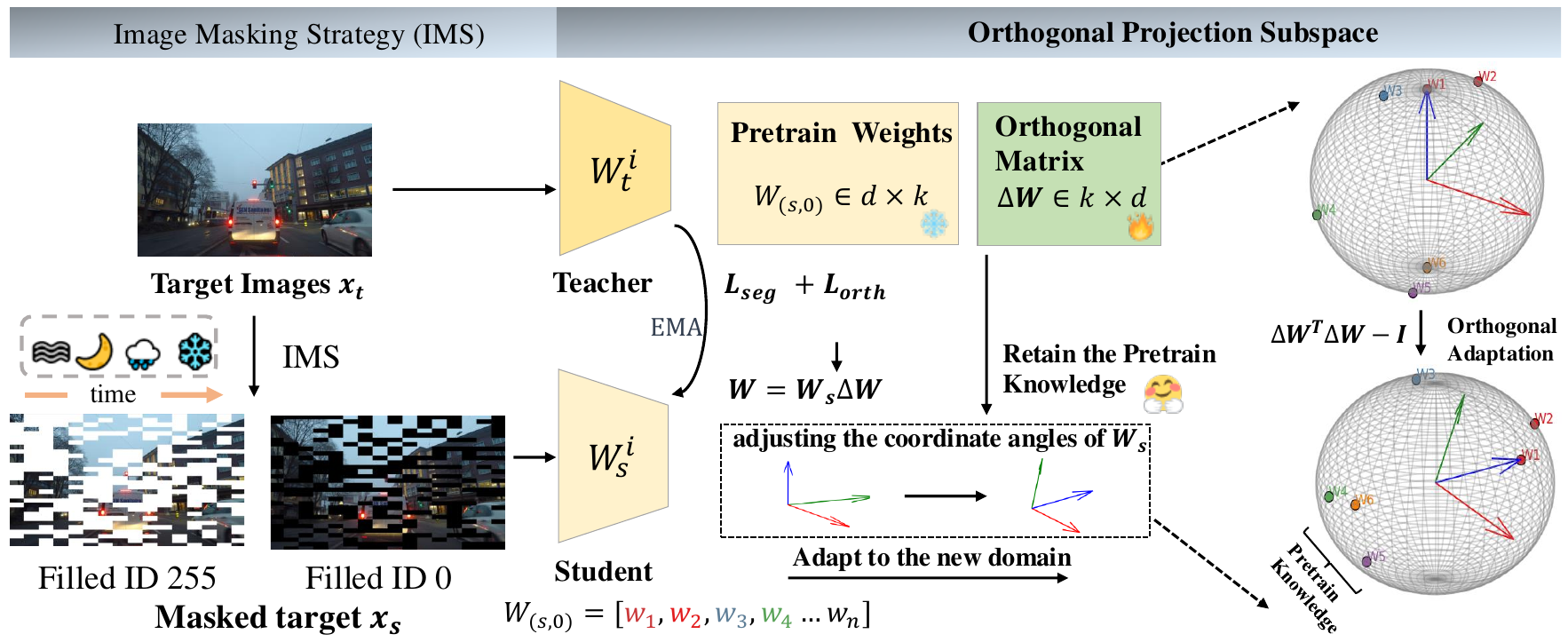}
\centering
\vspace{-0.20cm}
\caption{The overall paradigm of our method, \textbf{\textit{OoPk}}. Given unlabeled data from a new domain, its original and simulated version by our IMS for unseen domains are fed into the teacher and student models, respectively. The teacher model then supervises the student model's adaptation to the new domain using our proposed orthogonal projection subspace tuning. This method fixes the original model's weights $\bm{W}_{s}$, introduces orthogonal adaptation weights $\bm{\Delta W}$, and adjusts the weight space angles to adapt to the new domain. 
}
\label{fig:method}
\vspace{-0.4cm}
\end{figure*}

\subsection{Preliminaries}

\noindent \textbf{Continual Test Time Adaptation (CTTA)}~\cite{wang2022continual} is first proposed to adapt a pre-source model $\bm{f_s}$ to rapidly and continuously changing target domains. Suppose that a task sequence samples $\mathrm{\textbf{\textit{X}}}_{t} = \{ \mathrm{\textit{X}}^{1}_{t}, \mathrm{\textit{X}}^{2}_{t}, \dots, \mathrm{\textit{X}}^{c}_{t}, \dots \}.$ CTTA scenario consists of two basic assumptions: (1) the mode can not access the source dataset after deploying it to the test time. (2) the model adaptation has to be finalized online in an unsupervised manner, meaning that each sample within the given target domains can only be used once. In terms of semantic segmentation tasks, CTTA targets predicting a pre-defined category for each pixel through softmax $\hat{y_c} = f_c(\bm{x}_t^c)$ in the specific target domain $c$, where $\bm{x}_t^c$ is sampled from $\mathrm{\textit{X}}^{c}_{t}$. For simplicity, we present $\bm{x}$ as follows.

\noindent \textbf{Orthogonal Transformation} is defined for a real square matrix whose columns and rows are orthogonal vectors, so called orthogonal matrix. This can be expressed as:
\vspace{-0.15cm}
\begin{equation}
    Q^{T}Q=QQ^{T} = I, Q \in \mathbb{R}^{n \times n}
\vspace{-0.15cm}
\end{equation}
where $Q^{T}$ is the transpose of $Q$ and $I$ denotes the identity matrix. Typically, orthogonal regularization~\cite{brock2018large,vorontsov2017orthogonality} has been widely leveraged to apply a soft constraint regularization of the weight $W$ to keep it close to orthogonal:
\vspace{-0.15cm}
\begin{equation}
    \lambda \sum_{i} || W^{T}_{i}W_{i} - I||,
\vspace{-0.15cm}
\end{equation}
In this way, the regularization optimizes the weight matrix towards an orthogonal subspace to cover diverse optimization directions and capture comprehensive data distributions, while stabilizing the model training and enhancing generalizability, which is under-explored in TTA or CTTA scenarios. In Fig.~\ref{fig:motivation}, we conduct a toy experiment to demonstrate that fine-tuning the pre-trained weight directions orthogonally is rational to preserve the semantic information, leading to maintaining the knowledge integrity of the pre-trained model while allowing it to absorb new domain knowledge. To achieve this, we can rotate or reflect all the neurons that naturally introduce orthogonal transformation. For more details please refer to our supplementary material.

\subsection{Orthogonal Projection Subspace tuning}
In this paper, to fully unleash knowledge from pre-trained source model whilst embracing continuously various unseen domain knowledge during the CTTA process. We adopt the widely used PEFT technique, LoRA~\cite{houlsby2019parameter,hu2022lora,poth-etal-2023-adapters} to implement our projection tuning which exhibits both effectiveness and efficacy, as illustrated in Fig.~\ref{fig:method}. In particular, when adapting a pre-trained source model $\bm{f_s}$ to a specific target distribution, a pre-trained weight matrix $\bm{W}_0 \in \mathbb{R}^{d \times k}$ will be adhered with a low-rank decomposition $\bm{W}_{s} = \bm{W}_0 + \Delta\bm{W} = \bm{W}_0 + BA$, where $B \in \mathbb{R}^{d \times r}, A \in \mathbb{R}^{r \times k}$, and the specified rank $r \ll min(d, k)$. During the CTTA adaptation tuning, $\bm{W}_0$ is frozen, while $A$ and $B$ contain trainable parameters. For an input $x$ to compute the output $h=\bm{W}_0 x$, now the forward pass becomes:
\vspace{-0.2cm}
\begin{equation}
    h = \bm{W}_0 x + \Delta\bm{W}x = \bm{W}_0 x + BAx,
\vspace{-0.20cm}
\end{equation}
wherein a random Gaussian initialization is used for $A$ and zero initialization for B. Hence, the model adaptation tuning begins with zero for $\Delta\bm{W}$, meaning the model will be optimized from the pre-trained source point. To further employ soft orthogonal regularization, we then utilize the orthogonal constraint to $\Delta\bm{W}$ and thus yield:
\vspace{-0.2cm}
\begin{equation}
\begin{aligned}
    \mathcal{L}_{orth} & = \left\| (\Delta\bm{W})^{T} \Delta\bm{W}-\bm{I} \right\|_{2} \\ 
    & = \left\| (\bm{BA})^{T} (\bm{BA})-\bm{I} \right\|_{2},
\end{aligned}
\vspace{-0.20cm}
\end{equation}
where $\left\| \cdot \right\|_{2}$ can be implemented with \textit{l2} loss. This can be viewed as a soft orthogonal constraint induced into additional tunable modules while still allowing the model to make utility of the pre-trained source knowledge, which is intrinsically suitable to tackle the difficult \textit{catastrophic forgetting} under the CTTA scenario. After that, the weight of student model $\bm{W}_{s}$ will be used to update teacher model $\bm{W}_{t}$ via an EMA framework. In this way, we are capable of maintaining the knowledge integrity from previous domain adaptation and gradually \textit{inject} unknown target knowledge into projected subspaces to ameliorate the model's adaptability, leading to \textit{harmony in diversity}. Noting that we can seamlessly merge additionally trainable $A$ and $B$ by $\bm{W}^{'} = \bm{W}_0 + BA$ to single one matrix $\bm{W}^{'}$, sharing the same size with $\bm{W}_0$, which introduces no latency overhead during inference when deployed and no model structure modification. Moreover, previous prompt- and MoE-based methods~\cite{liu2024continual,yang2024exploring,lee2024becotta} still rely on heavy trainable parameters tuning or weights Stochastic Restoration to resolve catastrophic forgetting problem during the adaptation, however, ours does not need such tricks and only tune much fewer trainable parameters, demonstrating superiority.

\subsection{Image Masking Strategy}

Image masking strategy plays a crucial role in self-supervised learning~\cite{devlin2018bert,bao2021beit,he2022masked,xie2022simmim,gao2024mimic,hou2022milan}, exhibiting intriguing explorations on the learning of feature representation. 
We delve into the analysis that due to no access to previous source samples and online target samples once only under the CTTA task, the model can not fully exploit knowledge from the target distribution domain. Moreover, we empirically observe that rapid and continuous target distribution usually comes with serious occlusion, blur, explosion, and darkness, such as foggy, night, rainy, and snowy conditions, \textit{etc.} This inspires us to propose a simple yet effective strategy Image Masking strategy (IMS) to maskout image content to simulate challenging target distribution, which progressively yet aggressively enhances student's learning capacity based on orthogonal projection subspace and ameliorates teacher's adaptability via the EMA framework.

To employ our IMS strategy, given an image input $x \in \mathbb{R}^{B \times 3 \times H \times W}$ where $B$ is the batch size, $H$ and $W$ correspond to height and width, we first generate a mask as follows:
\vspace{-0.2cm}
\begin{equation}
\begin{aligned}
    mask = \Theta(s, s)
\end{aligned}
\vspace{-0.1cm}
\end{equation}
where $\Theta(\cdot)$ illustrates uniform distribution to sample values within interval $[0, 1)$, and $s$ defines detailed $grid\_size$ of generated the mask, as shown in Fig.~\ref{fig:method} in terms of masked target. Then we utilize thresholding operation with a given masking ratio $\alpha$ to shift the soft masking vector to binary one, by:
\vspace{-0.2cm}
\begin{equation}
\begin{aligned}
   \mathbb{I}_{\{mask \geq \alpha\}} &= \begin{cases}
      1, & \text{if} \quad {mask}_{i, j} \geq \alpha\\
      0, & \text{otherwise},
      \label{eq:if}
    \end{cases} 
\end{aligned}
\vspace{-0.2cm}
\end{equation}
to obtain masking vector $\mathbb{I}_{\{mask \geq \alpha\}}$ which will be resized to $(H, W)$ to match the same spatial size with $x$, and then multiplied with image $x$ to get maskout input for student $\bm{f}_{s}$:
\vspace{-0.2cm}
\begin{equation}
    x_{s} = \mathbb{I}_{\{mask \geq \alpha\}} \cdot x,
\end{equation}
whereas keeping $x_{t} = x$ for teacher model $\bm{f}_{t}$ to force the online student adaptation aggregate diverse target prior knowledge by mimicking the potentially encountering target distribution. This gradually enhances the model's adaptability and supports high-quality pseudo labels to guide the model training, which appropriately mitigates error accumulation for the CTTA task.

Interestingly, one can consider assigning maskout pixels in $x_{s}$ with either $0$ or $255$ proxy pixels to stimulate the underlying target domains for \textit{night} and \textit{snow} weather, respectively. We conduct comprehensive experiments to showcase the promising improvements in terms of our IMS strategy. Recently, Continual-MAE~\cite{liu2024continual} also explores a similar MAE strategy to handle CTTA task, however, mainly focuses on conventional reconstruction objective employed on HoG feature, demonstrating different motivations, even with complex uncertainty selection to sample out unreliable regions. In contrast, our IMS strategy only requires a simple yet effective random masking mechanism for the student's pixel input and adopts an aggressive masking to enforce the model to extract task-relevant target knowledge quickly, leading to reduce risk of error accumulation.


\subsection{Overall Optimization Objective}

Following preliminary works~\cite{wang2022continual,lee2024becotta,liu2023vida,yang2024exploring}, we adopt a teacher-student framework (EMA) to process the adaptation tuning. After computing the pseudo labels $\hat{y_t}$ from teacher $\bm{f}_{t}$ with parameters $\theta_{t}$ based on $x_{t}$, the segmentation loss is then computed by cross-entropy loss between $\bm{f}_s$ and $\bm{f}_t$:
\vspace{-0.2cm}
\begin{equation}
    \mathcal{L}_{seg} = -\sum_c \hat{y_t} \log  \hat{y_s},
\label{eq:segloss}
\vspace{-0.2cm}
\end{equation}
where $\hat{y_s}$ is the prediction from student based on maskout $x_s$, and $c$ denotes the overall classes. The total training loss $\mathcal{L}$ for student model is:
\vspace{-0.2cm}
\begin{equation}
    \mathcal{L} = \mathcal{L}_{seg} + \lambda \mathcal{L}_{orth},
\vspace{-0.2cm}
\end{equation}
where $\lambda$ serves as a balancing coefficient and $\mathcal{L}_{orth}$ is our proposed orthogonal constraint loss. 

\begin{figure}[t]
    \includegraphics[width=0.5\textwidth]{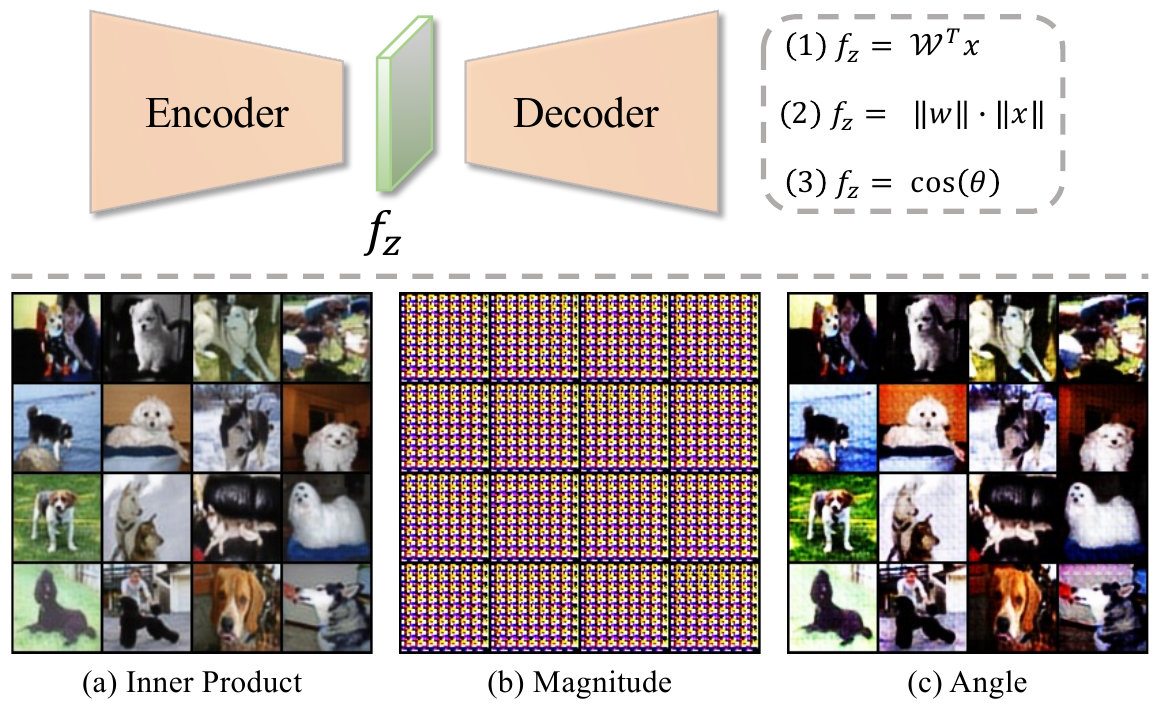}
    \centering
    \vspace{-0.7cm}
    \caption{A toy experiment to demonstrate the motivation of Orthogonal Projection Subspace tuning. The encoder-decoder is constructed and trained using inner product activation with \textit{MSE} loss, and (a) shows the final reconstruction. During testing, the angular information of weights alone can well reconstruct the input image, while magnitude can not. To preserve the semantic information of images, fine-tuning the neuron weight directions within the orthogonal manner will be rational.}
    \label{fig:motivation}
    \vspace{-0.4cm}
\end{figure}

Note that we only update the added tunable parameters while freezing the whole initial source model. After the update of the student model $\theta_{s} \to \theta^{'}_{s}$, we then update the teacher $\theta_{t}$ by moving average to gradually incorporate adaptability from student:
$\theta^{'}_{t} = \beta\theta_{t} + (1 - \beta)\theta^{'}_{s}$, where $\beta$ is a smoothing factor.

\section{Experiments}
\label{sec:method}

\noindent \textbf{Experimental settings and datasets.} Following the task setting outlined in~\cite{wang2022continual,lee2024becotta,liu2023vida,yang2024exploring}, CTTA is of the same setting as TTA but further sets the target domain rapidly and continuously changing, introducing more difficulties during the continual adaptation process. 

\noindent \textbf{Cityscapes-to-ACDC} is a continually changing semantic segmentation task designed to mimic continual distribution shifts in reality. Adverse Conditions Dataset (ACDC) shares the same classes with Cityscapes while containing various domain scenarios. It is collected in four different adverse visual conditions: Fog, Night, Rain, and Snow. For TTA validation, we use the pre-trained source model for each of the four ACDC target domains separately. For CTTA validation, we repeat the same sequence group (of the four conditional scenarios) 3 times (\textit{i.e.,} Fog $\to$ Night $\to$ Rain $\to$ Snow) in total 40 times to simulate real-world changes~\cite{wang2022continual}, 10 times referred to supplementary material.

\noindent \textbf{SHIFT} is a synthetic autonomous driving CTTA datasets~\cite{sun2022shift}. This dataset provides instead realistic domain shift along real-world environments, facilitating the development of CTTA methods, which includes a discrete set containing 4250 sequences and a continuous set with an additional 600 sequences. In this dataset for the CTTA task, the source model will be trained on discrete datasets and evaluated on the continuous validation dataset. The continuous set will be conducted as Daytime $\to$ Night, Clear $\to$ Foggy, and Clear $\to$ Rainy.

\noindent \textbf{Evaluation metrics.} We report all of the semantic segmentation results using mean-Intersection-over -Union (mIoU) in \%, along with pixel-level mean-Accuracy (mAcc) in \%.

\begin{table*}[htb]
\centering
\setlength\tabcolsep{11pt}
\begin{adjustbox}{width=1\linewidth,center=\linewidth}
\begin{tabular}{c|c|cc|cc|cc|cc|c|c }
\toprule
\hline

\multicolumn{2}{c|}{Test-Time Adaptation}          & \multicolumn{2}{c|}{Source2Fog}    & \multicolumn{2}{c|}{Source2Night}     & \multicolumn{2}{c|}{Source2Rain}  & \multicolumn{2}{c|}{Source2Snow}    & \multirow{2}{*}{Mean-mIoU$\uparrow$} & \multirow{2}{*}{Params$\downarrow$} \\ \cline{1-10}
Method &Venue &mIoU$\uparrow$ &mAcc$\uparrow$ 
&mIoU$\uparrow$ &mAcc$\uparrow$ &mIoU$\uparrow$ &mAcc$\uparrow$ &mIoU$\uparrow$ &mAcc$\uparrow$& \\ \hline
Source & NeurIPS2021 &69.1&79.4&40.3&55.6&59.7&74.4&57.8&69.9 &56.7 & / \\ 
TENT~\cite{wang2020tent}  & ICLR2021   &69.0&79.5&  40.3&55.5&  59.9&74.1&  57.7&69.7 &56.7 & 0.08M \\ 
CoTTA~\cite{wang2022continual} & CVPR2022 &70.9&80.2 &41.2&55.5 &62.6&75.4 &59.8&70.7&58.6 & 84.61M \\ 
DePT~\cite{gao2022visual} & ICLR2023  &71.0&80.2&40.9&\underline{55.8}&61.3&74.4&59.5&70.0&58.2 & N/A \\ 
VDP~\cite{gan2023decorate} & AAAI2023  &70.9&80.3&41.2&55.6&62.3&75.5&59.7&70.7&58.5 & N/A \\ 

SVDP~\cite{yang2024exploring} &AAAI2023 &72.1 &81.2 &\underline{42.0}&54.9& 64.4&\textbf{76.7} &\underline{62.2}&\textbf{72.8}&60.1 & $\textgreater$ 84.61M \\ 
\bottomrule
\rowcolor{tabhighlight} OoPk$_{r=32}$ &\textit{ours} &\textbf{73.6}&\textbf{81.8} &41.5&54.7& \underline{65.4}&\underline{76.0} &61.1&71.2&\underline{60.4} & 8.32M \\ 

\rowcolor{tabhighlight} OoPk$^{*}_{r=32}$ &\textit{ours} &\underline{73.5}&\underline{81.5} &\textbf{47.4}&\textbf{56.7}& \textbf{65.6}&75.9 &\textbf{63.0}&\underline{72.1}&\textbf{62.4} & 8.32M \\ 
\bottomrule
\end{tabular}
\end{adjustbox}
\vspace{-0.2cm}
\caption{Segmentation Performance comparison of Cityscapes-to-ACDC TTA. We use Cityscape as the source domain and ACDC as the four target domains in this setting. Mean-mIoU represents the average mIoU value in four TTA experiments. $OoPk^{*}$ denotes using data synthesis as BECoTTA~\cite{song2023ecotta} to warmup the added trainable parameters first and then conduct TTA. Note that we opt to compare with methods that released their corresponding codes in terms of TTA scenario here.
}
\label{tab:main_tta}
\vspace{-0.6cm}
\end{table*}

\noindent \textbf{Implementation details.} Our method \textit{OoPk} facilitates a flexible capacity, so it supports various variants according to the selection of rank $r$ and model architectures, CNNs, or transformers. For Cityscapes-to-ACDC, we employ Segformer-B5~\cite{xie2021segformer} pre-trained on Cityscapes as our off-the-shell source model. The resolution of $1920 \times 1080$ for ACDC images will be downsampled into $960 \times 540$. Following previous methods~\cite{wang2022continual,song2023ecotta,lee2024becotta,yang2024exploring}, multi-scale image factors $[0.5, 1.0, 1.5, 2.0]$ are applied to augment the teacher model, which is slightly different from previous work. For the SHIFT dataset, we follow the details in ~\cite{sun2022shift} to first pre-trained source model on discrete datasets and then conduct CTTA on the continual validation set. Both of them are evaluated predictions under the original resolution. We use Adam optimizer~\cite{kingma2014adam} with $(\beta_1, \beta_2) = (0.9, 0.999)$ with learning rate of 1e-4 and batch size of 1 for all experiments. All experiments are conducted on one NVIDIA A100 GPU and MMSeg~\cite{mmseg2020} using Pytorch~\cite{paszke2019pytorch}. We set our hyperparameters $r$ in OPS, IMS $grid\_size$ $s$, IMS masking ratio $\alpha$ ,and $\lambda$ with 32, 32, 0.75, and 1.0, respectively. 

\begin{table*}[htb]
\centering
\setlength\tabcolsep{2pt}
\begin{adjustbox}{width=1\linewidth,center=\linewidth}
\begin{tabular}{c|c|ccccc|ccccc|ccccc|c|c|c }
\toprule
\hline

\multicolumn{2}{c|}{Time}     & \multicolumn{15}{c}{$t$ \makebox[10cm]{\rightarrowfill} }                                                                              \\ \hline
\multicolumn{2}{c|}{Round}          & \multicolumn{5}{c|}{1}    & \multicolumn{5}{c|}{2}     & \multicolumn{5}{c|}{3}  & \multirow{2}{*}{Mean$\uparrow$}   & \multirow{2}{*}{Gain$\uparrow$} & \multirow{2}{*}{Params$\downarrow$}  \\ \cline{1-17}
Method & Venue & Fog & Night & Rain & Snow & Mean$\uparrow$ & Fog & Night & Rain & Snow  & Mean$\uparrow$ & Fog & Night & Rain & Snow & Mean$\uparrow$ & \\ \hline
Source & NeurIPS2021  &69.1&40.3&59.7&57.8&56.7&69.1&40.3&59.7& 	57.8&56.7&69.1&40.3&59.7& 57.8&56.7&56.7&/ & /\\ 
TENT~\cite{wang2020tent} & ICLR2021  &69.0&40.2&60.1&57.3&56.7&68.3&39.0&60.1& 	56.3&55.9&67.5&37.8&59.6&55.0&55.0&55.7&-1.0 & 0.08M \\ 
CoTTA~\cite{wang2022continual} & CVPR2022  &70.9&41.2&62.4&59.7&58.6&70.9&41.1&62.6& 	59.7&58.6&70.9&41.0&62.7&59.7&58.6&58.6&+1.9 & 84.61M \\ 
DePT~\cite{gao2022visual} & ICLR2023 
&71.0&40.8&58.2&56.8&56.5&68.2&40.0&55.4&53.7& 54.3&66.4&38.0&47.3&47.2&49.7&53.4&-3.3 & N/A \\
VDP~\cite{gan2023decorate} & AAAI2023  &70.5&41.1&62.1&59.5&  58.3    &70.4&41.1&62.2&59.4& 58.2     & 70.4&41.0&62.2&59.4& 58.2   &  58.2 & +1.5 & N/A \\
EcoTTA~\cite{song2023ecotta} & CVPR2023  &68.5 &35.8 &62.1 &57.4&  55.9    &68.3 &35.5 &62.3 &57.4 & 55.8     & 68.1 &35.3 &62.3 &57.3& 55.7   &  55.8 & -0.9 & 3.46M \\
SVDP~\cite{yang2024exploring} & AAAI2023 &72.1&44.0&65.2&63.0&61.1& 72.2&44.5&65.9&63.5&61.5  &72.1&44.2&65.6&63.6&61.4  &61.3 &+4.6 & $\textgreater$ 84.61M \\ 
C-MAE~\cite{liu2024continual} & CVPR2024  &71.9 &44.6 &67.4 &63.2 &61.8    &71.7 &44.9 &66.5 &63.1 &61.6     & 72.3 &45.4 &67.1 &63.1 &62.0    &  61.8 & +5.1 & $\textgreater$ 84.61M \\
 \bottomrule
BECoTTA$_{M}$~\cite{lee2024becotta} & ICML2024  &72.3 &42.0 &63.5 &59.8 & 59.4    &72.4 &41.9 &63.5 &60.2 & 59.5    & 72.3 & 41.9 & 63.6 & 60.2 & 59.5   &  59.5 & +2.8 & 2.15M\\
BECoTTA$^{*}_{M}~\cite{lee2024becotta}$ & ICML2024  & 71.8 & 48.0 & \underline{66.3} & 62.0 & 62.0 & 71.7 & 47.7 & 66.3 & 61.7 & 61.9 & 71.8 & 47.7 & 66.3 & 61.9 & 61.9 & 61.9 & +5.2 & 2.70M \\
BECoTTA$^{*}_{L}~\cite{lee2024becotta}$ & ICML2024  &72.7 & \underline{49.5} & \underline{66.3} & 63.1 & 62.9 & 72.6 & \underline{49.4} & 66.3 & 62.8 & 62.7 & 72.5 & 49.7 & 66.2 & 63.1 & 62.8& 63.0 & +6.3 & 11.86M\\
\bottomrule
  \rowcolor{tabhighlight} OoPk$_{r=4}$ & \textit{ours} & 73.1 & 42.3 & 65.5 & 61.5 & 60.6 & \underline{74.2} & 42.8 & 66.6 & 62.3 & 61.5 & \textbf{74.4} & 43.2 & \underline{66.4} & 63.0 & 61.8 & 61.3 & +4.6 & 1.04M \\ 
 \rowcolor{tabhighlight} OoPk$^{*}_{r=4}$ & \textit{ours} & \textbf{74.0} & 48.4 & 66.0 & \textbf{64.5} & \underline{63.2} & 73.9 & 49.2 & \underline{66.8} & \underline{64.1} & \underline{63.5} & 73.4 & \underline{50.0} & 66.3 & 63.3 & \underline{63.3} & \underline{63.3} & +\textbf{6.6} 
 & 1.04M \\ 
 \rowcolor{tabhighlight} OoPk$_{r=32}$ & \textit{ours} & 73.6 & 43.4 & 65.9 & 63.2 & 61.5 & \textbf{74.3} & 43.1 & \textbf{66.9} & \underline{64.1} & 62.1 & \textbf{74.4} & 42.8 & 66.2 & \underline{64.2} & 61.9 & 61.8 & +5.1 & 8.32M \\ 
 \rowcolor{tabhighlight} OoPk$^{*}_{r=32}$ & \textit{ours} & \underline{73.8} & \textbf{50.1} & \textbf{66.8} & \underline{64.1} & \textbf{63.7} & 73.4 & \textbf{50.3} & 66.7 & \textbf{65.0} & \textbf{63.8} & \underline{73.5} & \textbf{50.4} & \textbf{66.7} & \textbf{64.8} & \textbf{63.8} & \textbf{63.8} & +\textbf{7.1} 
 & 8.32M \\ 
 \bottomrule
\end{tabular}
\end{adjustbox}
\vspace{-0.2cm}
\caption{Performance comparison for Cityscapes-to-ACDC CTTA. We adopt Cityscape as the source domain and ACDC as the continual target domains. During testing, we sequentially evaluate the four target domains 3 times. Mean is the average score of mIoU. Gain refers to the improvement over the Source model, while Params means trainable parameters counting. $*$ denotes using data synthesis as BECoTTA~\cite{song2023ecotta} to warmup the added trainable parameters first and then conduct TTA.}
\label{tab:main_ctta}
\vspace{-0.4cm}
\end{table*}

\begin{figure*}[htb]
\centering
\includegraphics[width=0.99\linewidth]{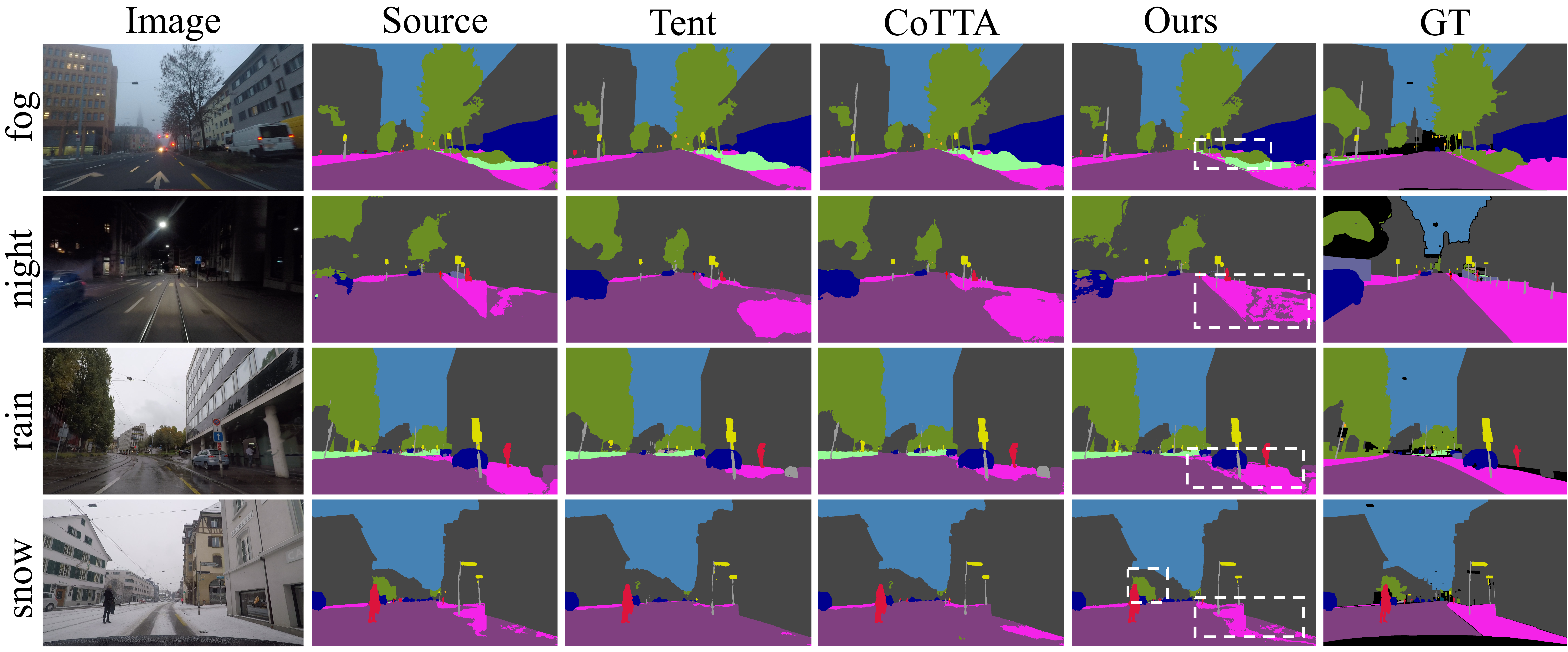}
\vspace{-0.2cm}
\caption{Qualitative comparison of our OoPk$_{r=32}$ (\textit{no} warmup) with previous SOTA methods on the ACDC dataset. Our method could better segment different pixel-wise classes such as shown in the white box. Best view zoom in and out.
}
\label{fig:vis}
\vspace{-0.6cm}
\end{figure*}

\begin{table}[t]
\centering
\setlength\tabcolsep{1pt}
\begin{adjustbox}{width=0.99\linewidth,center=\linewidth}
\begin{tabular}{c|cc|cc|cc|c }
\toprule
\hline
\multicolumn{1}{c|}{Scenarios}   & \multicolumn{2}{c|}{$Daytime - Night$}    & \multicolumn{2}{c|}{$Clear - Foggy$}     & \multicolumn{2}{c|}{$Clear - Rainy$}  &Mean    \\ \cline{1-7}
Method & mIoU & ACC & mIoU & ACC  & mIoU & ACC &mIoU $\uparrow$  \\ \hline
Source  & 73.54 & 77.00 & 50.83  &  52.31 & 79.12 & 83.31 & 67.83 \\ 
TENT~\cite{wang2020tent}  & 74.42 & 76.82  & 50.90 & 52.82  & 80.12 & 84.64 &  68.48\\
CoTTA~\cite{wang2022continual}  & 74.19 & 77.77 & \underline{52.08} & 53.08 & 79.85 & 84.51 & 68.70  \\
\hline
\rowcolor{tabhighlight} OoPk$_{r=4}$ & \textbf{75.05} & \textbf{78.49} &\underline{52.08} & \underline{53.68} & \underline{80.45} & \underline{84.9}&\underline{69.19} \\
\rowcolor{tabhighlight} OoPk$_{r=32}$ & \underline{75.04} & \underline{78.41} &\textbf{52.31} & \textbf{53.85} & \textbf{80.61} & \textbf{84.93}& \textbf{69.32}\\
 \bottomrule
 \end{tabular}
\end{adjustbox}
\vspace{-0.2cm}
\caption{Performance comparison for SHIFT dataset's continuous validation set. ACC denotes the average accuracy score.}
\vspace{-0.6cm}
\label{tab:main_ctta_shift}
\end{table}

\subsection{Comparison with Previous Methods} 
For the TTA scenario, as shown in Tab.~\ref{tab:main_tta}, directly employing the source model alone can get only 56.7\% mIoU across four domains. Recently, CoTTA and SVDP increase this to 58.6\% and 60.1\%, respectively. However, our method further improves performance to 60.4\%, while obtaining better 62.4\% when using data synthesis to warmup the added trainable parameters based on proposed OPS first and then conduct TTA, following BECoTTA~\cite{lee2024becotta}. Our method introduces no inference latency overhead using re-parameterization tricks when deploying, but SVDP, BECoTTA and DePT necessitate prompt or MoE vectors storage and computation either at the token-level or image-level, inducing extra overhead. Moreover, we only need to tune the added trainable modules and apply fewer multi-scale data-augmentations $[0.5, 1.0, 1.5, 2.0]$, different with CoTTA or SVDP $[0.5, 0.75, 1.0, 1.25, 1.5, 1.75, 2.0]$, all of which exhibit superior efficiency and effectiveness of \textit{OoPk}.

For the CTTA scenario, regarding the Cityscapes-to-ACDC setting, we can find in Tab.~\ref{tab:main_ctta} that our method obtains competitive performances when addressing continuously changing domain shifts, and achieves state-of-art when adopting a few steps warmup utilizing data synthesis based on source dataset Cityscapes raw image only, which has been demonstrated in BECoTTA~\cite{lee2024becotta} and we detail this procedure in supplementary material. Firstly, in comparison with BECoTTA and CoTTA, our OoPk$_{r=4}$ needs fewest 1.04M trainable parameters to realize 61.3\% and 63.3\% mIoU after adaptation tuning around 3 turns on ACDC val, which showcases the training efficacy of our method. When scaling up our trainable module by setting rank $r=32$, one can observe that our method consistently improves the performance to 61.8\% and 63.8\%, surpassing existing works. Interestingly, without warmup, our OoPk$_{r=4}$ with 1.04M params is able to outperform BECoTTA with much fewer trainable params, thanking to our proposed OPS tuning better attacking with catastrophic forgetting and IMS strategy reducing pseudo label error accumulation. Noting that BECoTTA establishes complicated MoE routing strategy and domain classification head to achieve adaptation, however, ours can be plugged into various model architectures without extra modification or inference deployment overhead.
We also observe our method is capable of gradually improving the same target domain performance under CTTA, such as 42.3\% $\to$ 42.8\% $\to$ 43.2\% at Night and 61.5\% $\to$ 62.3\% $\to$ 63.0\% at Snow for OoPk$_{r=4}$. This further demonstrates the effective adaptability of the cooperation from our proposed OPS and IMS, better addressing  catastrophic forgetting and error accumulation issues. Segmentation comparisons can be visualized in Fig.~\ref{fig:vis}.
Regarding the effectiveness of SHIFT, after pre-training on discrete datasets, we then conduct validation on continuous datasets, more pre-training details can refer to supplementary material. As shown in Tab.~\ref{tab:main_ctta_shift}, our method obtains 69.19\% with $r=4$ and 69.32\% with $r=32$, transcending all other methods. 
\begin{table*}[!ht]
    \centering
    \renewcommand{\arraystretch}{1.3}
    \setlength{\tabcolsep}{.33em}
    {\scriptsize
    \begin{tabular}{l|ccccccccccccccc|c}
    \hline
        Method & \rotatebox{90}{Gaussian}  & \rotatebox{90}{Shot} & \rotatebox{90}{Impulse} & \rotatebox{90}{Defocus} & \rotatebox{90}{Glass} & \rotatebox{90}{Motion} & \rotatebox{90}{Zoom} & \rotatebox{90}{Snow} & \rotatebox{90}{Frost} & \rotatebox{90}{Fog} & \rotatebox{90}{Brightness} & \rotatebox{90}{Contrast} & \rotatebox{90}{Elastic} & \rotatebox{90}{Pixelate} & \rotatebox{90}{Jpeg} & Avg. err $\downarrow$ \\ \hline
        Source & 80.1 & 77.0 & 76.4 & 59.9 & 77.6 & 64.2 & 59.3 & 64.8 & 71.3 & 78.3 & 48.1 & 83.4 & 65.8 & 80.4 & 59.2 & 69.7 \\ 
        TENT~\cite{wang2020tent} & 41.2 & 40.6 & 42.2 & 30.9 & 43.4 & 31.8 & 30.6 & 35.3 & 36.2 & 40.1 & 28.5 & 35.5 & 39.1 & 33.9 & 41.7 & 36.7 \\ 
        Continual TENT~\cite{wang2020tent} & 41.2 & 38.2 & 41.0 & 32.9 & 43.9 & 34.9 & 33.2 & 37.7 & 37.2 & 41.5 & 33.2 & 37.2 & 41.1 & 35.9 & 45.1 & 38.3 \\ 
        SWRNSP~\cite{choi2022improving} & 42.4 & 40.9 & 42.7 & 30.6 & 43.9 & 31.7 & 31.3 & 36.1 & 36.2 & 41.5 & 28.7 & 34.1 & 39.2 & 33.6 & 41.3 & 36.6 \\ 
        EATA~\cite{niu2022efficient} & 41.6 & 39.9 & 41.2 & 31.7 & 44.0 & 32.4 & 31.9 & 36.2 & 36.8 & 39.7 & 29.1 & 34.4 & 39.9 & 34.2 & 42.2 & 37.1 \\ 
        CoTTA~\cite{wang2022continual} & 43.5 & 41.7 & 43.7 & 32.2 & 43.7 & 32.8 & 32.2 & 38.5 & 37.6 & 45.9 & 29.0 & 38.1 & 39.2 & 33.8 & 39.4 & 38.1 \\ 
        EcoTTA~\cite{song2023ecotta} & 42.7 & 39.6 & 42.4 & 31.4 & 42.9 & 31.9 & 30.8 & 35.1 & 34.8 & 40.7 & 28.1 & 35.0 & 37.5 & 32.1 & 40.5 & 36.4 \\ 
        BECoTTA~\cite{lee2024becotta} (w/o SDA) & 42.1 & 38.0 & 42.2 & 30.2 & 42.9 & \textbf{31.7} & 29.8 & \textbf{35.1} & 33.9 & 38.5 & 27.9 & \textbf{32.0} & 36.7 & 31.6 & 39.9 & 35.5 \\
        \hline
        \rowcolor{tabhighlight} OoPk$^{*}_{r=4}$ (w/o warmup) & 41.8 & \textbf{37.4} & 42.6 & \textbf{30.0} & 43.2 & 32.3 & 29.9 & 35.8 & 33.2 & \textbf{38.1} & 28.7 & 33.0 & 36.8 & 32.8 & 40.3 & 35.7 \\
        \rowcolor{tabhighlight} OoPk$^{*}_{r=32}$ (w/o warmup) & \textbf{41.0} & 37.7 & \textbf{41.1} & 30.5 & \textbf{41.5} & 32.1 & \textbf{29.2} & \textbf{35.1} & \textbf{33.0} & 38.3 & \textbf{27.0} & 32.2 & \textbf{36.0} & \textbf{30.9} & \textbf{39.0} & \textbf{34.9} \\ \hline
    \end{tabular}
    }
    \vspace{-0.2cm}
    \caption{Classification error rate (\%) for CIFAR100-to-CIFAR100C. Results are evaluated on WideResNet-40. 
    }
    \vspace{-0.6cm}
\label{table:cifar100cc}
\end{table*}

In addition to evaluating our method on classification tasks, we provide results of CIFAR100-to-CICAR100C classifications to further demonstrate the generalizability of our OoPk.  As shown in Table.~\ref{table:cifar100cc}, OoPk$_r=32$ (w/o warmup) consistently outperforms existing methods, showing -3.2\% and -0.6\% reduction of the average error rate over CoTTA and BECoTTA, respectively. Though we observe a slightly higher error when using OoPk$_r=4$, our method still shows competitively superior -2.4\% to CoTTA.

\vspace{-0.1cm}
\subsection{Ablation Studies}
\vspace{-0.1cm}

To demonstrate the promising effectiveness and efficacy of each proposed component, we carry out various ablative experiments as shown in Tab.~\ref{tab:ablation_study}. Note that we perform ablations without any \textit{warmup}. When adopting the basic \textit{LoRA} method with $r=4$, the performance improves from 56.70\% mIoU to 58.63\%, but with a slight decrease in mAcc, which can be regarded as a basic adaptation tuning. By introducing the OPS to impose orthogonal constraint during adaptation for trainable parameters $\bm{\Delta W}$, we can further improve to 59.82\%, by 3.12\% beyond the source model. Then, we add our IMS strategy assigning masking pixels with 0, and the performance can be boosted to 61.30\% mIoU, which elucidates our IMS helps student model aggressively learning contextual relationships from target samples by masking input while gradually ameliorating teacher's adaptability, leading to overall improved performance. Besides, when setting masking pixels with 255 to stimulate explosion or snow conditions, we can still observe improvements, via utilizing masking pixels with 0 and 255 iteratively, the best performance of 61.50\% can be further acquired. 
Since this only brings marginal improvements, we opt for the simple IMS$_{pix0}$ only.  We also ablate the CTTA domain adaptation orders in Tab.~\ref{tab:ctta_order} by averaging 3 times, demonstrating no impact on CTTA orders. All of these results demonstrate the superiority of each component, OPS and IMS.

\begin{figure}[tbp]
    \begin{center}
    \centering 
\includegraphics[width=0.495\textwidth]{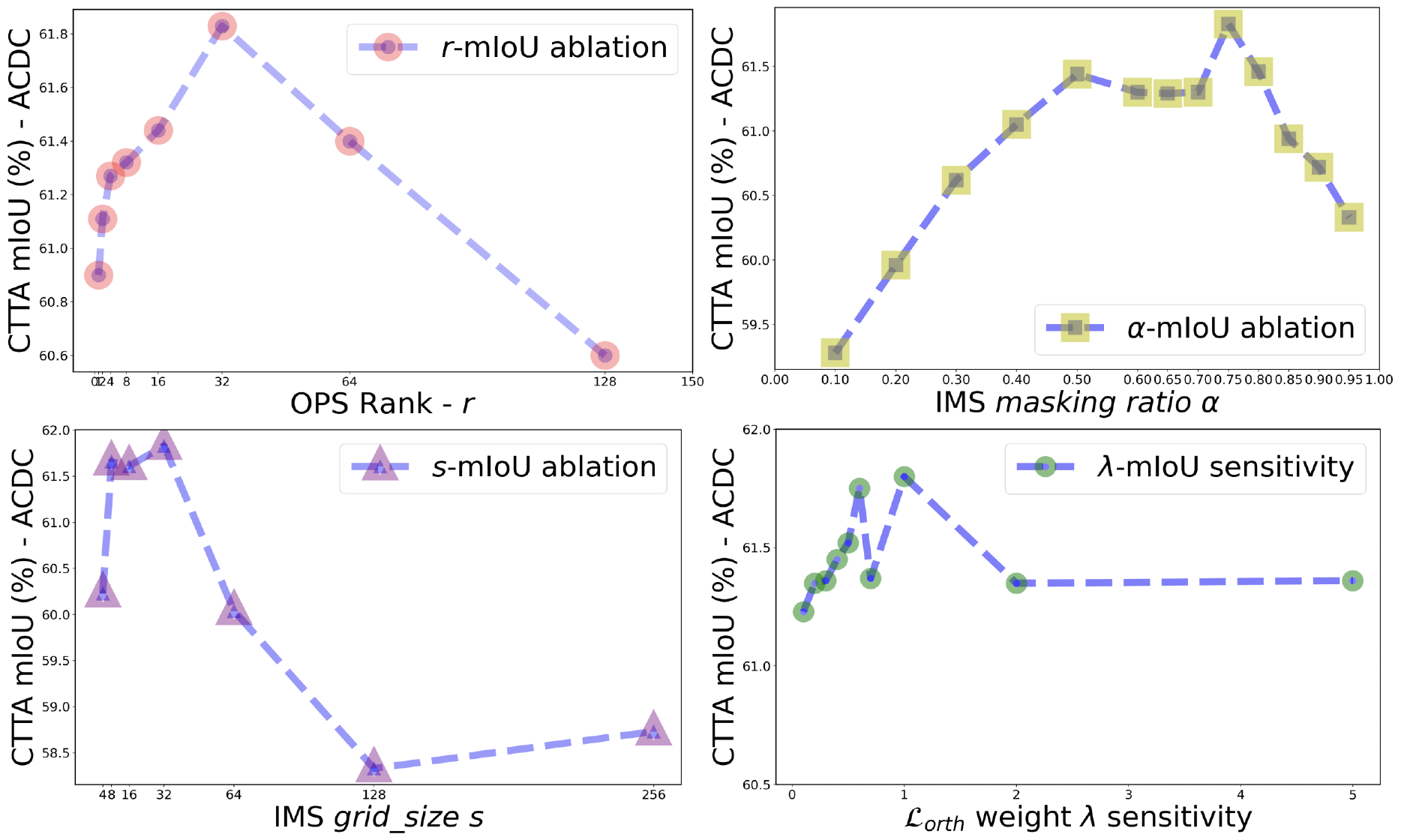}
    \end{center}
\setlength{\abovecaptionskip}{-0.05 cm}
\vspace{-0.2cm}
    \caption{Ablations about each component. We validate by gradually adding our proposed components to the overall training pipeline to demonstrate the improvement of each one.}
    \label{fig:ablations}
\vspace{-0.65cm}
\end{figure}

\noindent \textbf{Impact of rank choice in OPS.}
As shown in Fig.~\ref{fig:ablations}, we ablate the rank setting following the common usage in LoRA. We find that setting a rank with 32 strikes the best while a lower rank setting shows inferior performance and a much higher rank results in degradation. We deduce that lower-rank adaptation can not capture enough knowledge adaptation while much higher-rank will cause task-overfitting due to the limited online target samples used under the CTTA scenario. Note that this enables a flexible trade-off between performance and training costs among different scenarios.

\begin{table}[h]
    \vspace{-0.35cm}
    \centering
    \tabcolsep 3pt
    \scalebox{0.99}{
    \begin{tabular}{c|cccc}
    \hline
    {CTTA Order} & {F-N-R-S} & {N-R-S-F} & {R-S-F-N} & {S-F-N-R} \\
    \hline
    \textbf{avg mIoU (\%)} & 61.3 & 61.0 & 61.1 & 61.1 \\
    \textbf{avg mAcc (\%)}  & 71.0 & 70.9 & 70.9 & 71.0 \\ 
    \hline
    \end{tabular}
    }
    \vspace{-0.2cm}
    \caption{Ablations about changing the CTTA target domain adaptation orders, F:Fog, N:Night, R:Rain and S:Snow.}
    \label{tab:ctta_order}
    \vspace{-0.45cm}
\end{table}

\noindent \textbf{Impact of masking ration and masking grid size in IMS.}
In the IMS study, shown in Fig.~\ref{fig:ablations}, we ablate the masking size first , and can be found that an appropriate grid size $s=32$ leads to the highest performance while large ones degrade the overall performance drastically. We attribute this to larger grid size constructs much larger connected regions within masked pixels , and it makes the model difficult to seek out any local details to output reasonable predictions. We provide more analysis about this in the supplementary material.

\noindent \textbf{Sensitivity of loss weight $\lambda$.}
Since we devise an orthogonal regularization into our adaptation tuning procedure, the sensitivity of $\lambda$ balancing between $\mathcal{L}_{seg}$ and $\mathcal{L}_{orth}$ can be found in Fig.~\ref{fig:ablations}. The ablative studies showcase the overall performance is insensitive to $\lambda$, setting with default 1.0 slightly leads to the best performance.

\begin{table}[!tb]
\label{ablationDAP}
\centering
\setlength\tabcolsep{4.0pt}
\renewcommand\arraystretch{1}
\begin{tabular}{cccc|cc}
\toprule
\makecell*[c]{\textit{LoRA}} & \makecell*[c]{$\mathcal{L}_{orth}$} & \makecell*[c]{IMS$_{pix0}$} & \makecell*[c]{IMS$_{pix255}$}  & mIoU$\uparrow$ & mAcc 
 $\uparrow$\\
 \midrule
& & & & 56.70 & 69.82 \\ 
\checkmark &  & &  & 58.63 & 68.55\\
\checkmark  & \checkmark &  & & 59.82 & 70.03\\
\checkmark  &  & \checkmark & & 59.87 & 71.15\\
\checkmark & \checkmark &\checkmark  &  & 61.30 & 70.99\\
\checkmark & \checkmark & &\checkmark & 61.44 & 71.44\\
\checkmark &  \checkmark &\checkmark  &\checkmark & 61.50 & 71.24\\
\bottomrule
\end{tabular}
\vspace{-0.3cm}
\caption{Ablation: Improvement of each component. 
}
\vspace{-0.7cm}
\label{tab:ablation_study}
\end{table}

\section{Conclusion}
\label{sec:conclusion}

In this paper, to better tackle catastrophic forgetting, we propose a novel Orthogonal Projection Subspace tuning with orthogonal regularization, which allows the model to adapt to new changing domains while also preserving the knowledge integrity of the previous domains. Furthermore, a simple yet effective Image Masking Strategy is proposed to mimic encountering scenarios to aggressively enforce student model learn richer target knowledge to enhance adaptability, while ameliorating the teacher's knowledge and reducing error accumulation. Our method obtains competitive even SoTA performances for TTA and CTTA scenarios in complex semantic segmentation or classification tasks, which hopes shed new light on future works.



{
    \small
    \bibliographystyle{ieeenat_fullname}
    \bibliography{main}
}

\clearpage
\maketitlesupplementary

\begin{abstract}
This supplementary material provides additional details and explanations to support our main paper, as follows:
\begin{itemize}
    \item Sec.~\ref{sec:orth_motivation} Toy Experiment to Clarify Motivation of Orthogonal Projection Subspace tuning.
    \item Sec.~\ref{sec:ortho_position_ablation} Additional Ablations of Orthogonal Projection Subspace tuning Position.
    \item Sec.~\ref{sec:add_rst} Additional Results of 10 Rounds.
    \item Sec.~\ref{sec:warmup_imple} Warmup Implementations.
    \item Sec.~\ref{sec:IMS_imple} Additional Image Masking Strategy Analysis.
    \item Sec.~\ref{sec:ims_b} Image Masking Strategy to BECoTTA.
    \item Sec.~\ref{sec:time_b} Training/Inference time compared with BECoTTA.
    \item Sec.~\ref{sec:shift_imple} SHIFT Discrete Pre-training Details.
    \item Sec.~\ref{sec:limitation} Limitation and Future Works.
\end{itemize}
\end{abstract}

\section{Toy Experiment to Clarify Motivation of Orthogonal Projection Subspace tuning}
\label{sec:orth_motivation}
In this section, we make more explanations for the motivation for the proposed Orthogonal Projection Subspace. First, based on previous studies~\cite{chen2020angular,liu2018decoupled,liu2017deep}, the angles of weights in neural networks capture most informative characteristics. Then we conduct toy experiments as illustrated in Fig.~\ref{fig:motivation} to showcase.
For the architecture of the toy experiment encoder-decoder, we apply three convolution neural layers in the encoder, and each layer adopts a downsampling scale to reduce the spatial resolution of the feature map and upscale the corresponding dimensions. After that, three deconvolution neural layers are employed to upsample the spatial resolution of the feature map to map with the input size in the decoder. To obtain the results shown in Fig.~\ref{fig:motivation}, we utilize Stanford Dogs dataset~\cite{khosla2011novel} to conduct the training following the standard inner product to produce the feature map $z$ by the convolution layer weight $w$ and input $x$ with inner product manner. During the testing, we apply $z=||w|| \cdot ||x||$ as the magnitude to obtain the reconstructed result in Fig.~\ref{fig:motivation} (b) while using $z=cos(\theta)$ where $\theta$ denotes the angle between the weight $w$ and the feature map $x$ to compute the result in Fig.~\ref{fig:motivation} (c). The toy experiment is completed with a learning rate of 0.001 in Adam optimizer~\cite{kingma2014adam} for 200 epochs.

The results in Fig.~\ref{fig:motivation} demonstrate that the angular information of weights can nearly reconstruct the input images, whereas the magnitude of weights lacks meaningful information. Noting that, the cosine activation in Fig.~\ref{fig:motivation} (c) is not used during training; instead, training relies solely on the inner product. These findings suggest that the angles (or directions) of weights are primarily responsible for encoding the informative characteristics of the input images. Consequently, fine-tuning the directions of weights is likely to be an effective approach for adopting the pre-trained weights for the new domain knowledge covering more comprehensive distributions while preserving the knowledge integrity of the pre-trained source model to alleviate catastrophic forgetting. Therefore, fine-tuning the directions of weights is to uniformly rotate or reflect all neuron weights within the same layer, inherently introducing an orthogonal transformation. 

\begin{figure}[t]
    \includegraphics[width=0.5\textwidth]{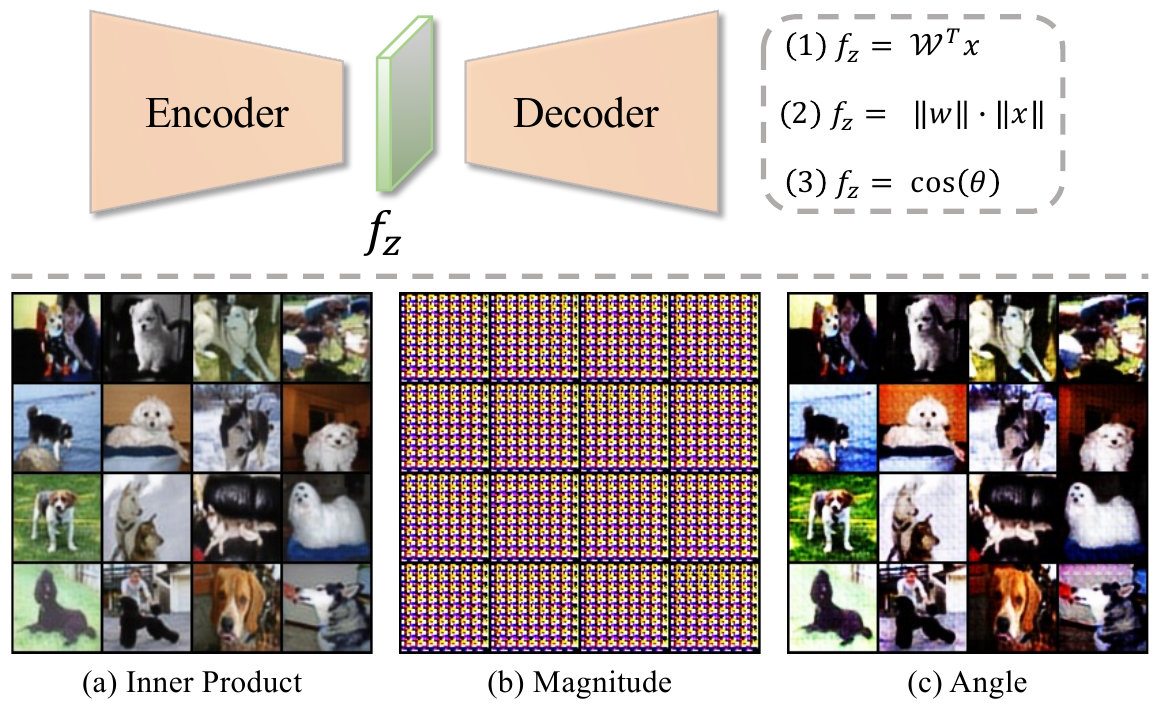}
    \centering
    \caption{A toy experiment to demonstrate the motivation of Orthogonal Projection Subspace tuning. The encoder-decoder is constructed and trained using inner product activation with \textit{MSE} loss, and (a) shows the final reconstruction. During testing, the angular information of weights alone can well reconstruct the input image, while magnitude can not. To preserve the semantic information of images, fine-tuning the neuron weight directions within the orthogonal manner will be rational.}
    \label{fig:motivation}
\end{figure}

\section{Additional Ablations of Orthogonal Projection Subspace tuning Position}
\label{sec:ortho_position_ablation}

In this section, we ablate the proposed Orthogonal Projection Subspace tuning (OPS) in various positions given the source pre-trained model, SegFormer B5, including both Attention\&FFN, Attention and FFN shown in Fig.~\ref{fig:param_compaison}. We can observe that equipping both Attention and FFN blocks in the source model with OPS achieves the best performance at 61.3\% mIoU on ACDC val. In contrast, all the other two variants, Attention, and FFN can also be superior to BECoTTA and CoTTA under \textit{no} warmup. Moreover, all our variants need much fewer trainable parameters than CoTTA and BECoTTA. In our main paper, we set the default with the Attention\&FFN variant. All of these further showcase the effectiveness and efficacy of our method.

\section{Additional Results of 10 Rounds}
\label{sec:add_rst}
\begin{figure*}[htb]
\centering
\includegraphics[width=0.99\linewidth]{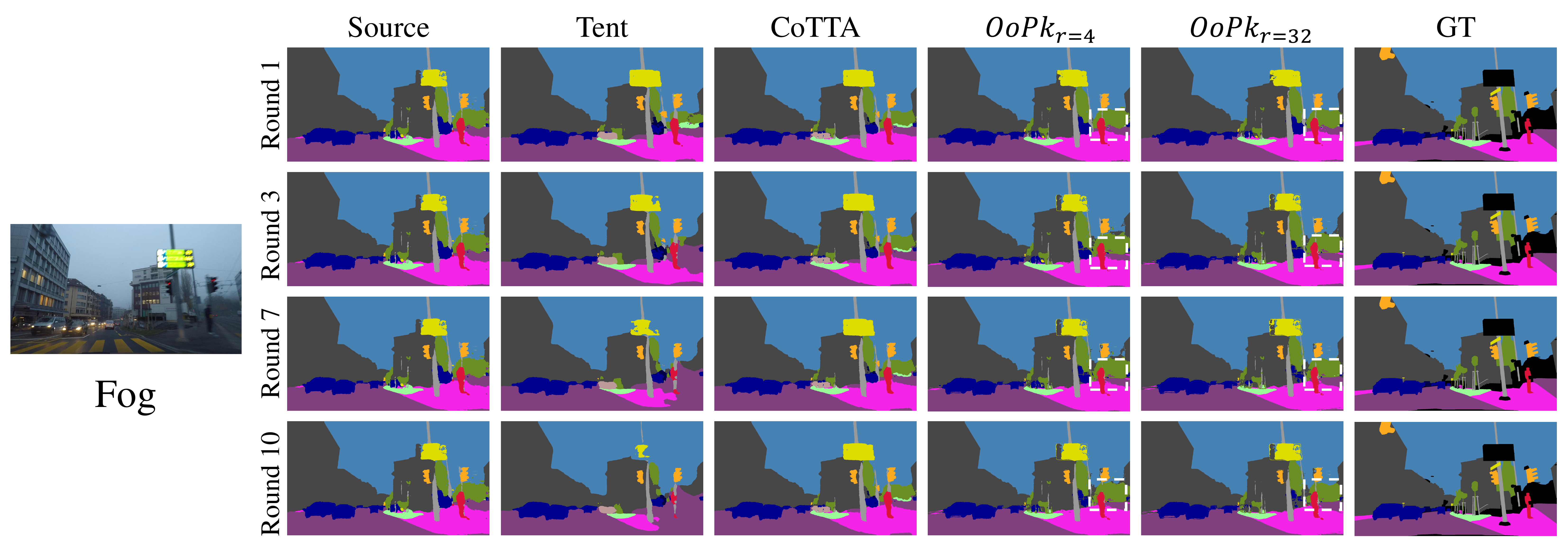}
\caption{Qualitative comparison of our method in foggy weather on ACDC val.
}
\label{fig:vis_fog}
\end{figure*}

\begin{figure*}[htb]
\centering
\includegraphics[width=0.99\linewidth]{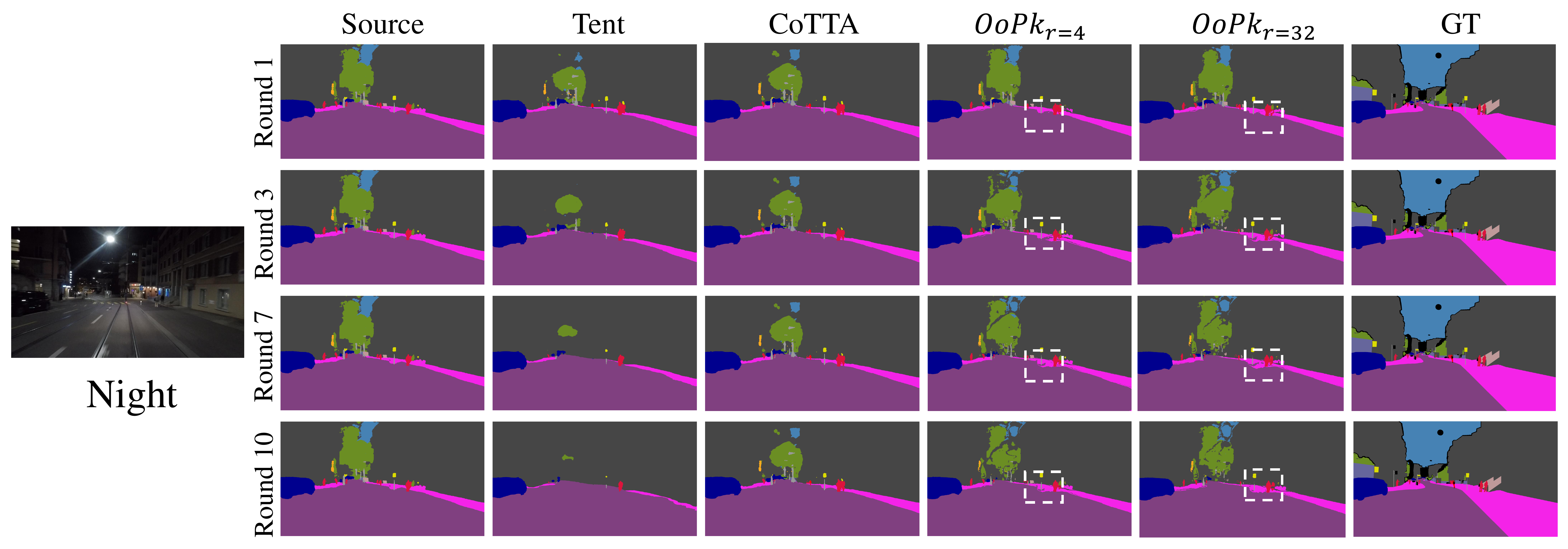}
\caption{Qualitative comparison of our method in night weather on ACDC val.
}
\label{fig:vis_night}
\vspace{6cm}
\end{figure*}

\begin{figure*}[htb]
\centering
\includegraphics[width=0.99\linewidth]{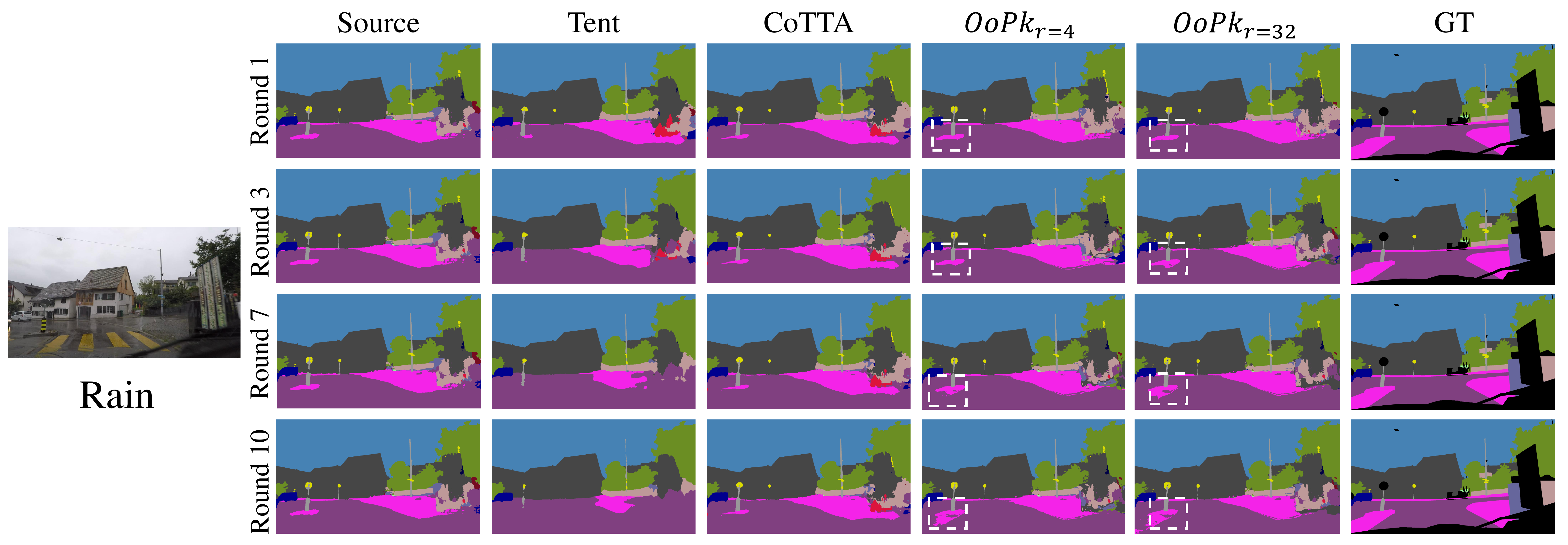}
\caption{Qualitative comparison of our method in rainy weather on ACDC val.
}
\label{fig:vis_rain}
\end{figure*}

\begin{figure*}[htb]
\centering
\includegraphics[width=0.99\linewidth]{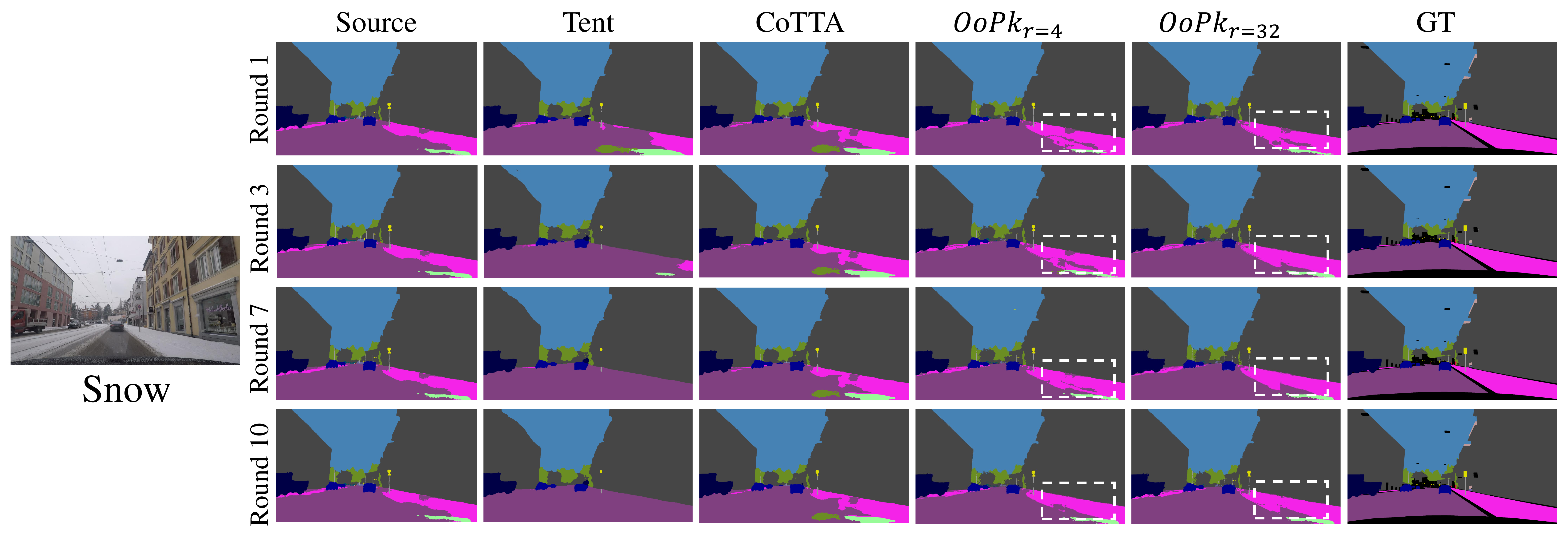}
\caption{Qualitative comparison of our method in snowy weather on ACDC val.
}
\label{fig:vis_snow}
\vspace{6cm}
\end{figure*}

In this section, we provide more comparative results in terms of 10 rounds on the ACDC dataset. As shown in Tab.~\ref{table:main1_ten}, our $OoPk_{r=4}$ can obtain 61.2\% and 62.9\% mIoU for \textit{no} warmup and warmup settings, which surpasses $BECoTTA_{M}$ and $BECoTTA^{*}_{M}$ and presents fewer trainable parameters. Moreover, BECoTTA, SVDP, or C-MAE still rely on Stochastic Restoration like CoTTA to mitigate catastrophic forgetting issue during the long-term adaptation, but ours does not thanks to orthogonal projection subspace which allows the model to adapt to new domains while preserving the knowledge integrity of the pre-trained source model to naturally alleviate catastrophic forgetting.
This further demonstrates that our overall approach is able to better attack catastrophic forgetting and error accumulation issues. We illustrate more segmentation results in Fig.~\ref{fig:vis_fog} for foggy conditions, in Fig.~\ref{fig:vis_night} for night conditions, in Fig.~\ref{fig:vis_rain} for rainy conditions, and in Fig.~\ref{fig:vis_snow} for snowy conditions, respectively.

Here we also introduce comparative results with recent ViDA~\cite{liu2023vida}, C-MAE~\cite{liu2024continual} and SVDP~\cite{yang2024exploring} in Tab.~\ref{table:main1_ten}. Note that, we pinpoint that both ViDA and C-MAE have not yet released the segmentation part codes, therefore we opt for the reported results in their paper and conjecture some implementations from their classification codes to deduce the necessary training parameters counting. As shown in  Tab.~\ref{table:main1_ten} and Tab. 2\&3 in our main paper, we can observe that our method still obtains competitive performance without \textit{warmup} and surpasses these competitive works when applying \textit{warmup} under much fewer trainable parameters scenario. However, ViDA, SVDP, and C-MAE allow the whole pre-trained source model weights training which is similar to CoTTA, leading to heavy training overhead of more than 84.6M training parameters that hampers the potential practical deployment in reality. For details about ViDA, one can refer to: training params collection\footnote{training params: \url{https://github.com/Yangsenqiao/vida/blob/main/cifar/vida.py##L162}} and training params in optimizer\footnote{training params in optimizer: \url{https://github.com/Yangsenqiao/vida/blob/main/cifar/cifar100c_vit.py##L175}}. For details about C-MAE, one can refer to: training params collection\footnote{training params: \url{https://github.com/RanXu2000/continual-mae/blob/master/cifar/vida.py##L162}} and training params in optimizer\footnote{training params in optimizer: \url{https://github.com/RanXu2000/continual-mae/blob/master/cifar/cifar10c_vit_mae.py##L228}}. For details about SVDP, one refer to: training params collection\footnote{training params: \url{https://github.com/Anonymous-012/SVDP/blob/main/mmseg/apis/test.py##L115}} and training params in optimizer\footnote{training params in optimizer: \url{https://github.com/Anonymous-012/SVDP/blob/main/mmseg/apis/test.py##L122}}.

\section{Warmup Implementations}
\label{sec:warmup_imple}

\subsection{Warmup data synthesis}
\begin{figure*}[htb]
\centering
\includegraphics[width=0.99\linewidth]{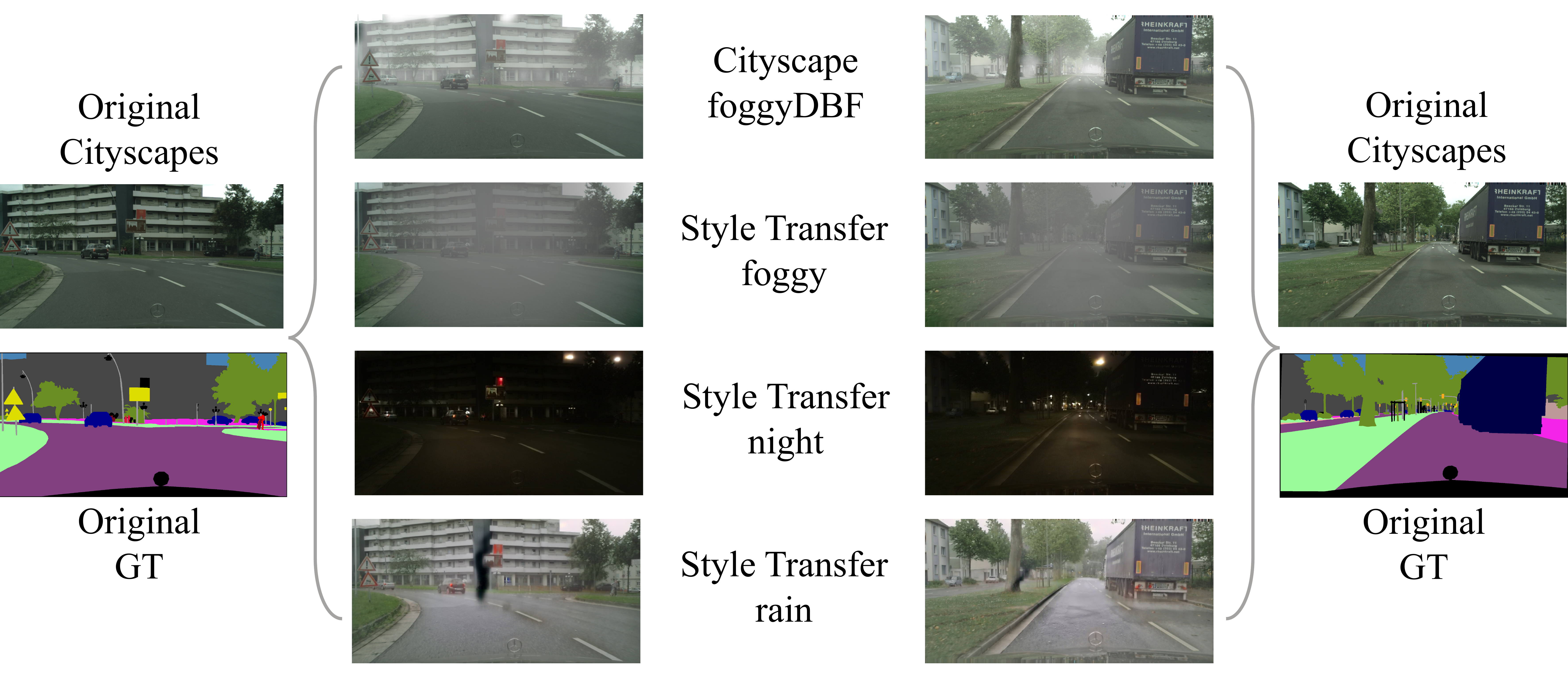}
\caption{Visualization of our synthetic data generation.
}
\label{fig:vis_dataAug}
\end{figure*}

\begin{figure*}[htb]
\centering
\includegraphics[width=0.99\linewidth]{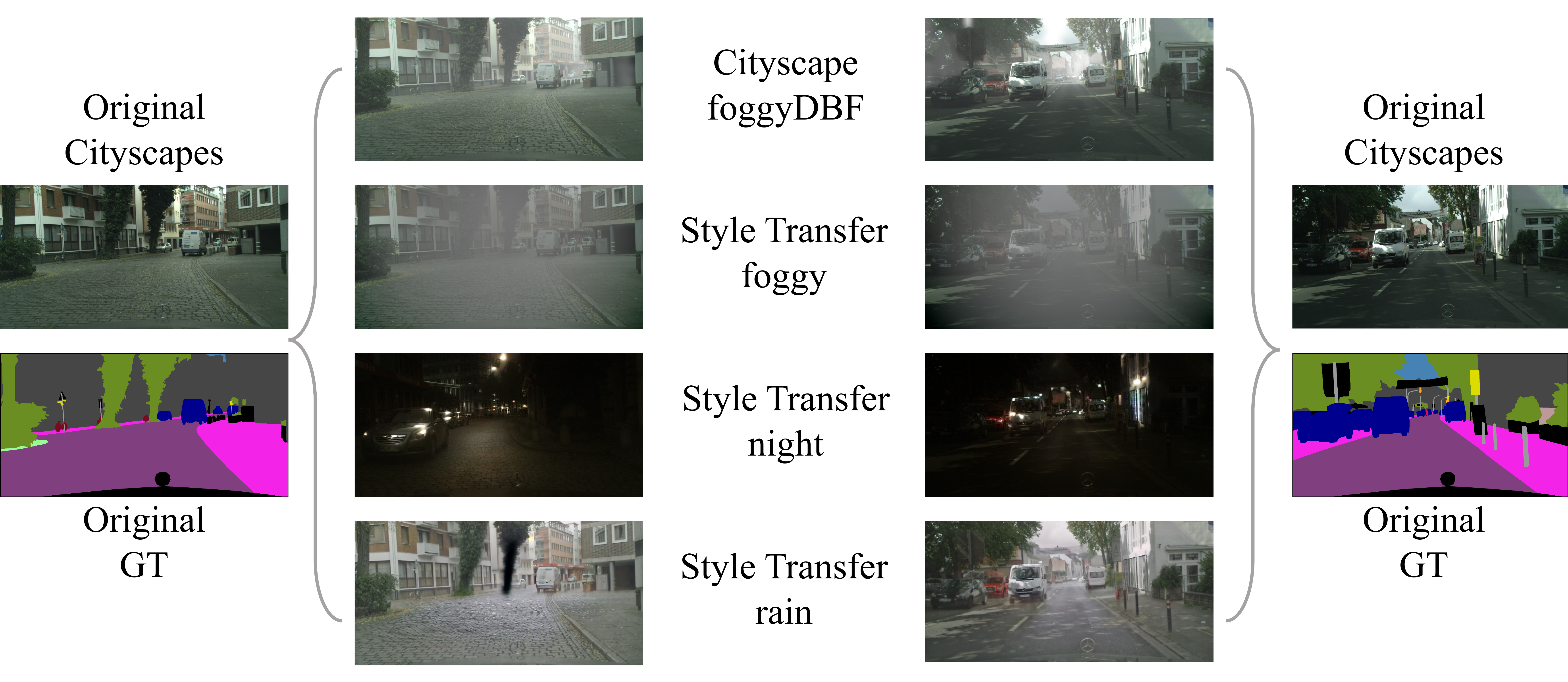}
\caption{Visualization of our synthetic data generation.
}
\label{fig:vis_dataAug2}
\end{figure*}

Following previous works, like BECoTTA~\cite{lee2024becotta}, we utilize pre-trained style transformer ~\cite{parmar2024one,jiang2020tsit} to first generate potential domains using Cityscapes RGB images only. To achieve this objective, we set the candidate domains with foggy, night, and rainy styles to simulate the real-world weather changes as demonstrated in Fig.~\ref{fig:vis_dataAug} and Fig.~\ref{fig:vis_dataAug2}. Note that, this is different from BECoTTA which adopts both TSIT~\cite{jiang2020tsit} and Pytorch augmentation to construct more aggressive data synthesis. And we will release our synthetic data to foster future work as BEcoTTA has not yet released this part.

\subsection{Warmup training details}
Here we display our warmup implemental details in Tab.~\ref{table:ours_params}.

\section{Additional Image Masking Strategy Analysis}
\label{sec:IMS_imple}
\begin{figure*}[htb]
\centering
\includegraphics[width=0.99\linewidth]{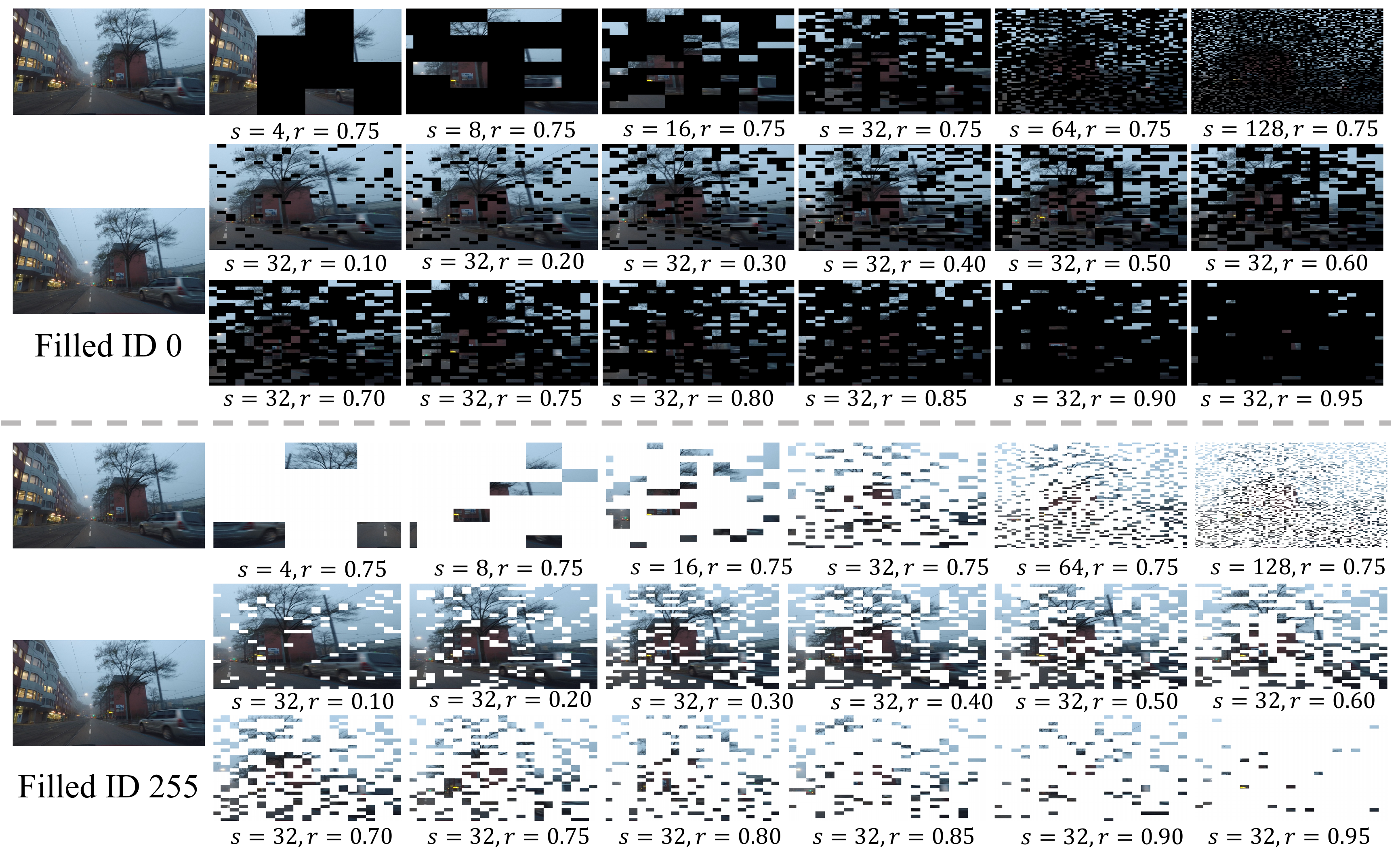}
\caption{Image Masking Strategy Variants visualizations.
}
\label{fig:vis_masked}
\vspace{6cm}
\end{figure*}

In this section, we visualize our Image Masking Strategy (IMS) to compare various masking implementations. In our main paper, we ablate the experimental results with different $grid\_size$ $s$ and masking ratio $\alpha$. From Fig.~\ref{fig:vis_masked}, we can see that a lower masking ratio or smaller $grid\_size$ can not maskout enough regions to enforce the student model better adapt to target domains while higher masking ratio or larger $grid\_size$ will cause the maskout regions nearly losing all potential contextual information. Only an appropriate masking ratio $\alpha$ and $grid\_size$ can obtain competitive performance, realizing strong adaptability learning for the student model and then gradually ameliorating the teacher's knowledge, when trying to mimic the potential target distributions to fill the masked pixels with 0 or 255. We also want to make differences with recent C-MAE~\cite{liu2024continual} which employs a making strategy on the input but targets at reconstructing the HoG feature while complex uncertainty selection is needed to sample out unreliable regions. However, our proposed IMS strategy only necessitates a simple yet effective random masking mechanism for student input images without challenging reconstruction objectives which we surmise will confuse the high-level segmentation or classification tasks. In this way, the proposed IMS enforces the model to gradually extract task-relevant target knowledge to reduce the risk of error accumulation and work cooperatively with Orthogonal Projection Subspace tuning to better tackle the error accumulation and catastrophic forgetting, respectively.

Regarding additional evaluation on the classification task on CIFAR100-CIFAR100C,~\cite{krizhevsky2009learning}, we utilize the backbone WideResNet-40 for CIFAR100CIFAR100C or a fair comparison with other methods.

\begin{table}[h]
        \centering
        
        \renewcommand{\arraystretch}{1.2}
        {\footnotesize
            \begin{tabular}{l|cc}
            \hline
            \multicolumn{1}{c|}{} & Warmup & TTA \\ \hline
            Dataset & Synthetic Data & Target domains \\
            Optimizer & AdamW & Adam \\
            Optimizer momentum & \multicolumn{2}{c}{\small{$(\beta_1, \beta_2)=(0.9, 0.999)$}} \\
            Iterations & 20K & Online \\
            Batch size & \multicolumn{2}{c}{1} \\
            Learning rate & 0.00006 & 0.0006/8 \\
            Label accessibility & Yes & No \\ \hline
            \end{tabular}%
        }%
        \caption{Our method hyperparameters.}
        \label{table:ours_params}
        \vspace{-.5cm}
     
\end{table}

\begin{table*}[t]
\vspace{2cm}
\centering
\renewcommand{\arraystretch}{1.3}
\setlength{\tabcolsep}{.23em}
\resizebox{0.99\linewidth}{!}{%
\begin{tabular}{lc|cccccccccccccccc|c|c}
\hline
\multicolumn{2}{l|}{\textbf{Round}} & \multicolumn{4}{c|}{\textbf{1}} & \multicolumn{4}{c|}{\textbf{3}} & \multicolumn{4}{c|}{\textbf{7}} &  \multicolumn{4}{c|}{\textbf{10}} & \multicolumn{1}{l}{} & \multicolumn{1}{l}{} \\ \hline

\multicolumn{1}{l|}{Method} & Venue & Fog & Night & Rain & \multicolumn{1}{c|}{Snow} 
& Fog & Night & Rain & \multicolumn{1}{c|}{Snow} 
& Fog & Night & Rain & \multicolumn{1}{c|}{Snow} 
& Fog & Night & Rain & Snow & Mean & Params$\downarrow$ \\ \hline

\multicolumn{1}{l|}{Source only} & \textit{NIPS'21} & 69.1 & 40.3 & 59.7 & \multicolumn{1}{c|}{57.8} 
& 69.1 & 40.3 & 59.7 & \multicolumn{1}{c|}{57.8} 
& 69.1 & 40.3 & 59.7 & \multicolumn{1}{c|}{57.8} 
& 69.1 & 40.3 & 59.7 & 57.8 & 56.7 & / \\

\multicolumn{1}{l|}{BN Stats Adapt~\cite{nado2020evaluating}} & \textit{-} & 62.3 & 38.0 & 54.6 & \multicolumn{1}{c|}{53.0} 
& 62.3 & 38.0 & 54.6 & \multicolumn{1}{c|}{53.0} 
& 62.3 & 38.0 & 54.6 & \multicolumn{1}{c|}{53.0} 
& 62.3 & 38.0 & 54.6 & 53.0 & 52.0 & 0.08M \\ 

\multicolumn{1}{l|}{Continual TENT~\cite{wang2020tent}} & \textit{ICLR'21} & 69.0 & 40.2 & 60.1 & \multicolumn{1}{c|}{57.3} 
& 68.3 & 39.0 & 60.1 & \multicolumn{1}{c|}{56.3} 
& 64.2 & 32.8 & 55.3 & \multicolumn{1}{c|}{50.9}     
& 61.8 & 29.8 & 51.9 & 47.8 & 52.3 & 0.08M \\                

\multicolumn{1}{l|}{CoTTA~\cite{wang2022continual}} & \textit{CVPR'22} 
& 70.9 & 41.2 & 62.4 & \multicolumn{1}{c|}{59.7} 
& 70.9 & 41.0 & 62.7 & \multicolumn{1}{c|}{59.7}      
& 70.9 & 41.0 & 62.8 & \multicolumn{1}{c|}{59.7}    
& 70.8 & 41.0 & 62.8 & 59.7 & 58.6 & 84.61M \\

\multicolumn{1}{l|}{SAR~\cite{niu2023towards}} & \textit{ICLR'23} 
& 69.0 & 40.2 & 60.1 & \multicolumn{1}{c|}{57.3} 
& 69.1 & 40.3 & 60.0 & \multicolumn{1}{c|}{57.8} 
& 69.1 & 40.2 & 60.3 & \multicolumn{1}{c|}{57.9} 
& 69.1 & 40.1 & 60.5 & 57.9 & 56.8 & 0.08M \\

\multicolumn{1}{l|}{ViDA~\cite{liu2023vida}} & \textit{ICLR'24} 
& 71.6 & 43.2 & \underline{66.0} & \multicolumn{1}{c|}{\underline{63.4}} 
& 73.2 & 44.6 & \textbf{67.2} & \multicolumn{1}{c|}{\textbf{64.2}} 
& 72.3 & 44.8 & \textbf{66.5} & \multicolumn{1}{c|}{62.9} 
& 72.2 & \underline{45.2} & \textbf{66.5} & 62.9 & 61.6 & $\textgreater$84.61M \\

\multicolumn{1}{l|}{EcoTTA~\cite{song2023ecotta}} & \textit{CVPR'23} 
& 68.5 & 35.8 & 62.1 & \multicolumn{1}{c|}{57.4} 
& 68.1 & 35.3 & 62.3 & \multicolumn{1}{c|}{57.3} 
& 67.2 & 34.2 & 62.0 & \multicolumn{1}{c|}{56.9} 
& 66.4 & 33.2 & 61.3 & 56.3 & 55.2 & 3.46M \\ 

\multicolumn{1}{l|}{BECoTTA$_{M}$~\cite{lee2024becotta}} & \textit{ICML'24} 
& 72.3 & 42.0 & 63.5 & \multicolumn{1}{c|}{60.1} 
& 72.3 & 41.9 & 63.6 & \multicolumn{1}{c|}{60.2} 
& 72.3 & 41.9 & 63.6 & \multicolumn{1}{c|}{60.3} 
& 72.3 & 41.9 & 63.5 & 60.2 & 59.4 & 2.15M \\

\multicolumn{1}{l|}{BECoTTA$^{*}_{M}$~\cite{lee2024becotta}} & \textit{ICML'24} 
& 71.8 & \underline{48.0} & \textbf{66.3} & \multicolumn{1}{c|}{62.0} 
& 71.8 & \underline{47.7} & 66.3 & \multicolumn{1}{c|}{61.9} 
& 71.8 & \underline{47.8} & \underline{66.4} & \multicolumn{1}{c|}{61.9} 
& 71.8 & \textbf{47.9} & \underline{66.3} & 62.6 & \underline{62.0} & 2.70M \\

\rowcolor{tabhighlight} \multicolumn{1}{l|}{OoPk$_{r=4}$} & \textit{Ours} 
& \underline{73.1} & 42.3 & 65.5 & \multicolumn{1}{c|}{61.5} 
& \textbf{74.4} & 43.2 & \underline{66.4} & \multicolumn{1}{c|}{63.0} 
& \textbf{73.8} & 42.4 & 65.5 & \multicolumn{1}{c|}{\textbf{63.8}} 
& \textbf{73.6} & 42.0 & 64.1 & \underline{63.5} & 61.2 & 1.04M \\

\rowcolor{tabhighlight} \multicolumn{1}{l|}{OoPk$^{*}_{r=4}$} & \textit{Ours} 
& \textbf{74.0} & \textbf{48.4} & \underline{66.0} & \multicolumn{1}{c|}{\textbf{64.5}} 
& \underline{73.4} & \textbf{50.0} & 66.3 & \multicolumn{1}{c|}{\underline{63.3}} 
& \underline{73.2} & \textbf{48.0} & \textbf{66.5} & \multicolumn{1}{c|}{\underline{63.6}} 
& \underline{73.1} & \textbf{47.9} & \textbf{66.5} & \textbf{63.6} & \textbf{62.9} & 8.32M \\

\hline

\end{tabular}
}
\caption{Performance for 10 rounds comparisons on ACDC val set. * denotes adopting warmup strategy as BECoTTA~\cite{lee2024becotta}.}
\label{table:main1_ten}
\vspace{-1.2cm}
\end{table*}

\begin{figure*}[htb]
\vspace{1cm}
\centering
\includegraphics[width=0.99\linewidth]{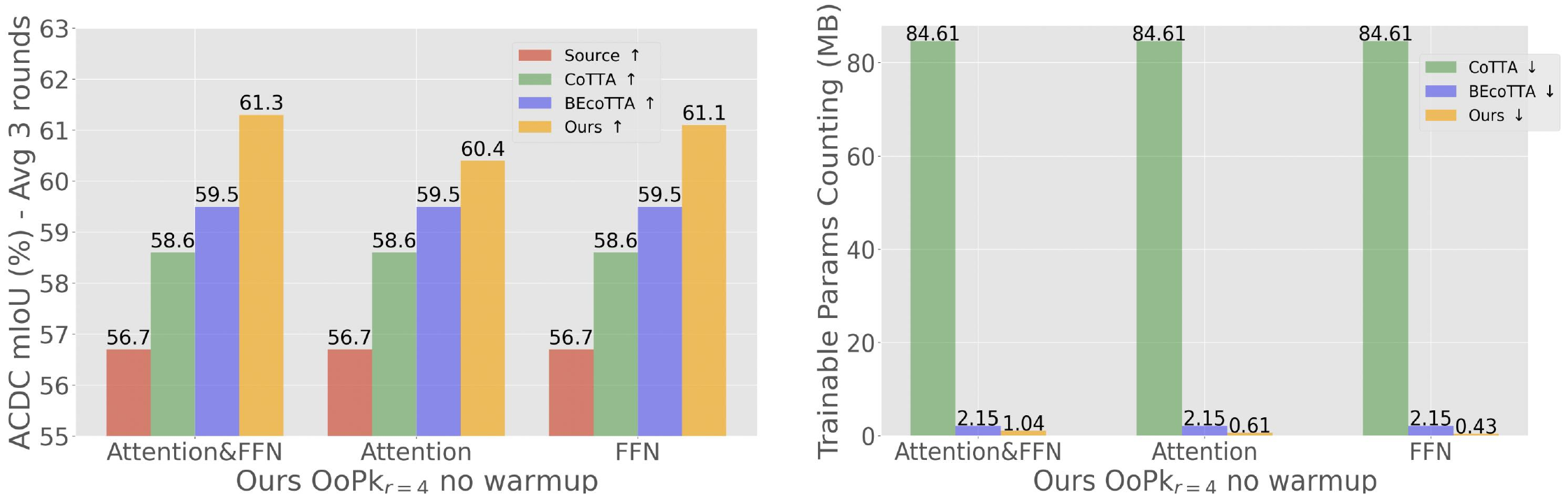}
\vspace{-0.2cm}
\caption{Comparisons on ablation for various OPS positions equipped into source model and corresponding tunable parameters counting. 
}
\label{fig:param_compaison}
\vspace{-0.4cm}
\end{figure*}

\begin{figure*}[htb]
\centering
\includegraphics[width=0.99\linewidth]{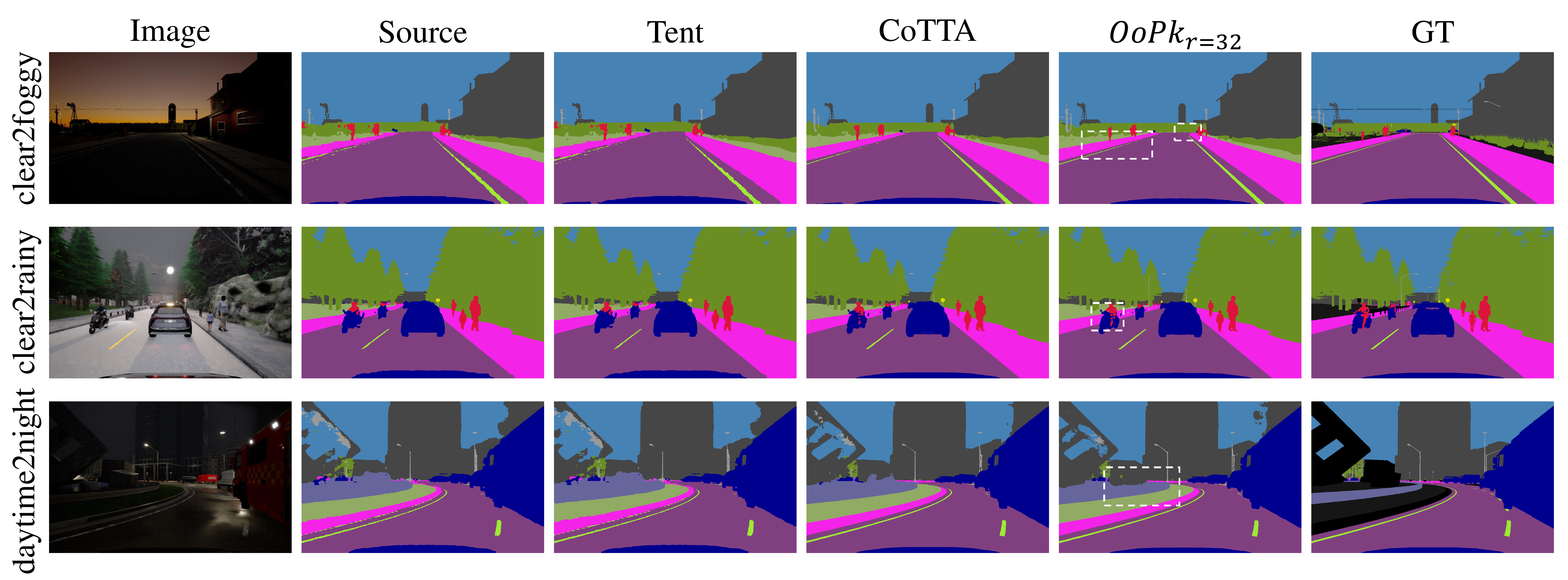}
\caption{Qualitative comparison of our method on SHIFT continuous val.
}
\label{fig:vis_shift_seg}
\end{figure*}

\section{Image Masking Strategy to BECoTTA}
\label{sec:ims_b}
To validate that our OPS and IMS work cooperatively, we then conduct comparative experiments that employ our proposed IMS to BECoTTA, reimple-BECoTTA$_{M}$ and BECoTTA$_{M}$ \textit{w} masking. We obtain 59.6\% mIoU and 59.7\% mIoU for reimple-BECoTTA$_{M}$ and BECoTTA$_{M}$ \textit{w} masking, respectively. We posit this is due to the primary MoE routing structure in BECoTTA while detailed \textit{BECoTTA\_style} or MoE\_style image masking strategies need to be designed to obtain improvements. This further demonstrates that our proposed OPS first projects an orthogonal subspace to acquire new target domain knowledge, while an aggressive image masking strategy enforces the model to adapt to target domains within the projected subspace quickly, leading to overall competitive adaptability.

\section{Training/Inference time compared with BECoTTA}
\label{sec:time_b}
While our method requires much fewer tunable parameters than BECoTTA, BECoTTA 2.15M \textit{v.s.} ours 1.04M, We further conduct training time comparisons, obtaining~1.38~hours for BECoTTA$_{M}$,~1.01~hours for our OoPk$_{r=4}$ and~1.02~hours for OoPk$_{r=32}$, respectively, which still demonstrates more efficiency of our method. We adopt the same comparative environment with one single Nvidia A6000 48GB.
Since our method can adopt a reparameterization trick to achieve the same inference latency with the source model which we consider not fair for BECoTTA as the MoE routing model cannot while BECoTTA also induces extra domain-aware classification heads.

\section{SHIFT implemental details}
\label{sec:shift_imple}

To conduct experiments on the SHIFT dataset, we first pre-train the source model on discrete datasets. Our implemental hyper-parameters are shown in Tab:~\ref{table:ours_shift_params}.

\vspace{-0.2cm}
\begin{table}[h]
        \centering
        
        \renewcommand{\arraystretch}{1.2}
        {\footnotesize
            \begin{tabular}{l|c}
            \hline
            \multicolumn{1}{c|}{} & pre-train \\ \hline
            Dataset & Discrete SHIFT Data \\
            Optimizer & AdamW \\
            Optimizer momentum & \multicolumn{1}{c}{\small{$(\beta_1, \beta_2)=(0.9, 0.999)$}} \\
            Iterations & 80K \\
            Batch size & \multicolumn{1}{c}{8} \\
            Learning rate & 0.00006 \\
            Label accessibility & Yes \\ \hline
            \end{tabular}%
        }%
        \caption{Our method hyperparameters.}
        \label{table:ours_shift_params}
     
\vspace{-0.7cm}
\end{table}

\section{Limitation and Future Works}
\label{sec:limitation}
We verify our OoPk demonstrates superior performance with fewer parameters compared to other CTTA baselines. However, it requires detailed hyperparameters choices, including the rank $r$ in OPS, the masking $grid\_size$ $s$ and masking ratio $\alpha$ in IMS, and loss weight tuning $\lambda$ in $\mathcal{L}_{orth}$. All of these will lead to slight performance fluctuations while also being considered as the flexibility of our method. Based on all our experimental results, we have demonstrated the superiority and efficacy that our OoPk consistently outperforms all current CTTA baselines which not only establishes competitive performances but also needs fewer trainable parameters and induces no inference overhead when deploying, regardless of the hyperparameter settings. We delve into different ways to tackle catastrophic forgetting and error accumulation issues in the CTTA scenario. For future works, since there are very few benchmarks for CTTA dense prediction tasks, including semantic segmentation, object detection, \textit{etc.} Meanwhile, unifying various modalities into this scenarios~\cite{li2022expansion,li2022weakly,li2025cross,liu2024less,zhao2025fishertune,li2023weakly,xu20243d,zang2024generalized,zeng2022s2,zhang2024mpt,luo2024mmevol} will be an interesting explorations, going beyond 2D and extending to 3D domains. We believe this will be meaningful for the community to develop more comprehensive and suitable benchmarks to flourish this community.

\clearpage



\end{document}